%% file: ms.tex
\documentclass[letterpaper, 10 pt, conference]{ieeeconf}  

\IEEEoverridecommandlockouts

\usepackage{graphics} 
\usepackage{epsfig} 
\usepackage{times} 
\usepackage{amsmath} 
\usepackage{amssymb}  

\usepackage{graphicx}
\usepackage{booktabs}

\include{chapters/tikz_commands}
\include{chapters/commands}

\usepackage[pagebackref,breaklinks,colorlinks]{hyperref}

\usepackage[capitalize]{cleveref}
\crefname{section}{Sec.}{Secs.}
\Crefname{section}{Section}{Sections}
\Crefname{table}{Table}{Tables}
\crefname{table}{Tab.}{Tabs.}

\title{\LARGE \bf
	RECALL: Rehearsal-free Continual Learning for Object Classification
}

\author{Markus Knauer$^{1}$, Maximilian Denninger$^{1,2}$ and Rudolph Triebel$^{1,2}$
	
	\thanks{$^{1}$ Institute of Robotics and Mechatronics, German Aerospace Center (DLR), Oberpfaffenhofen, Germany
		{\tt\small {first}.{last}@dlr.de}}%
	\thanks{$^{2}$ Technical University of Munich (TUM), Germany
		{\tt\small triebel@in.tum.de}
	}
}

\begin{document}

\maketitle
\thispagestyle{empty}
\pagestyle{empty}

\begin{abstract}
\input{chapters/abstract}
\end{abstract}

\section{Introduction}
\input{chapters/introduction}

\section{Related work}\label{relatedWork}
\input{chapters/relatedWork}

\section{\OURAPPROACH}
\input{chapters/approach}

\section{\DLRDataset dataset}
\input{chapters/howsDataset}

\section{Results} \label{sec_evaluation}
\input{chapters/evaluation}

\section{Conclusion}
\input{chapters/conclusion}

\small{
	\bibliographystyle{IEEEtran}
	\bibliography{IEEEabrv,ms.bib}
}
\end{document}


\title{- Supplemental Material - \\RECALL: Rehearsal-free Continual Learning for Object Classification}
%
%

\maketitle

\section{Introduction}

With \textit{\OURAPPROACH}, we introduce a novel algorithm for rehearsal-free object classification on 2D images for continual learning.
This supplemental document includes the quantitative results of the graphs, which are shown in the result section of the paper.
For more details, see \cref{exactResults}.
In \cref{eval_seq}, a study on catastrophic forgetting is provided.
Furthermore, we give a more detailed overview of our \DLRDataset dataset in \cref{dataset}.
More ablation study results are presented in \cref{AddAblationStudies}.
And finally, we also discuss the release of our used hyperparameters in \cref{sourceCode}.
In addition to this document, a video is provided, making the approach more accessible.
For more details for downloading \DLRDataset and running the code, check the accompanying code and README file.

\section{Quantitative results}\label{exactResults}

In this section, we provide the quantitative results of our experiments on the different datasets from Fig.\ 5 of the paper.
\Cref{tab_comp_core50} shows the accuracy of each tested approach on the \coreFifty dataset.
In \cref{tab_comp_cifar100}, the results of the iCIFAR dataset are printed.
The result of the different approaches on our \DLRDataset dataset are shown in \cref{tab_comp_hows}. 
Finally, the long version of the \DLRDataset dataset is depicted in \cref{tab_comp_hows_long}.

\section{Qualitative sequence specific results}\label{eval_seq}
In the paper, we examine the average performance of the variants of \OURAPPROACH over all sequences. 
In this section, we want to further investigate the performance for the categories of each sequence in each sequence.
We therefore provide plots for all four different versions of \OURAPPROACH in combination with \coreFifty (\cref{fig:seq_eval_core50}), \iCIFAR (\cref{fig:seq_eval_icifar}), \DLRDataset (\cref{fig:seq_eval_hows}), and the long version of the \DLRDataset dataset (\cref{fig:seq_eval_hows_long}).
In each plot, it can be seen how the categories of a particular sequence perform in the following sequences.
For example, the categories' performance of the first sequence is shown in red during the training.
In the second sequence, the green categories are added and tracked throughout the training process.
The same is true for the remaining sequences.

\input{tikz_figures/MyDatasetApple}

\begin{figure*}
	\plotSequenceFig{Plots/Sequence_evaluation/core50_Recall.csv}{8}{0}{\coreFifty with \OURAPPROACH}{\coreFifty}{fig:coreSeqRecall} 
	\hfill
	\plotSequenceFig{Plots/Sequence_evaluation/core50_Recall_reg.csv}{8}{0}{\coreFifty with \Recallreg}{\coreFifty}{fig:coreSeqRecallReg}
	
	\plotSequenceFig{Plots/Sequence_evaluation/core50_Recall_var.csv}{8}{0}{\coreFifty with \Recallvar}{\coreFifty}{fig:coreSeqRecallVar}
	\hfill
	\plotSequenceFig{Plots/Sequence_evaluation/core50_Recall_var_reg.csv}{8}{0}{\coreFifty with \Recallvarreg}{\coreFifty}{fig:coreSeqRecallVarReg}
	\caption{The results on the \coreFifty dataset with our four methods.}
	\label{fig:seq_eval_core50}
\end{figure*}

\begin{table*}
	\centering 
	\caption{Comparison of \OURAPPROACH and other approaches on \coreFifty dataset, shown in Fig. 5 of the paper}
	\label{tab_comp_core50}	
	\begin{tabular}{|l|c|c|c|c|c|c|c|c|c|}
		\hline
		Approach & Seq $0$ & Seq $1$ & Seq $2$ & Seq $3$ & Seq $4$ & Seq $5$ & Seq $6$ & Seq $7$ & Seq $8$ \\ \hline 
		LwF & $97.85$ & $85.44$ & $77.22$ & $68.23$ & $58.97$ & $50.78$ & $44.46$ & $38.14$ & $34.14$ \\ \hline 
		EWC & $97.64$ & $75.93$ & $69.85$ & $64.93$ & $57.66$ & $52.07$ & $48.63$ & $46.99$ & $43.29$ \\ \hline 
		SI & $97.71$ & $75.95$ & $62.20$ & $47.79$ & $42.33$ & $36.41$ & $32.76$ & $29.65$ & $26.77$ \\ \hline 
		AR1 & $95.01$ & $84.48$ & $76.89$ & $72.95$ & $65.26$ & $61.37$ & $59.52$ & $58.59$ & $57.67$ \\ \hline 
		AR1 from [18] & $97.34$ & $87.64$ & $84.62$ & $80.84$ & $78.23$ & $74.71$ & $72.24$ & $70.85$ & $69.48$ \\ \hline
		\OURAPPROACH & $97.78$ & $91.11$ & $86.15$ & $78.94$ & $72.98$ & $68.19$ & $64.76$ & $66.22$ & $67.01$ \\ \hline
		\Recallvar & $97.56$ & $93.0$ & $86.22$ & $75.0$ & $66.67$ & $60.44$ & $51.81$ & $51.53$ & $53.01$  \\ \hline
		\Recallreg & $97.78$ & $90.67$ & $82.52$ & $79.11$ & $74.49$ & $71.15$ & $68.44$ & $67.44$ & $67.33$ \\ \hline
		\Recallvarreg & $96.89$ & $90.44$ & $86.44$ & $81.61$ & $79.47$ & $75.52$ & $73.68$ & $72.97$ & $\bm{72.02}$ \\ \hline
	\end{tabular}
\end{table*}

\begin{figure*}
	\plotSequenceFig{Plots/Sequence_evaluation/cifar100_Recall.csv}{9}{0}{\iCIFAR with \OURAPPROACH}{\iCIFAR}{fig:icifarSeqRecall} 
	\hfill
	\plotSequenceFig{Plots/Sequence_evaluation/cifar100_Recall_reg.csv}{9}{0}{\iCIFAR with \Recallreg}{\iCIFAR}{fig:icifarSeqRecallReg}
	
	\plotSequenceFig{Plots/Sequence_evaluation/cifar100_Recall_var.csv}{9}{0}{\iCIFAR with \Recallvar}{\iCIFAR}{fig:icifarSeqRecallVar}
	\hfill
	\plotSequenceFig{Plots/Sequence_evaluation/cifar100_Recall_var_reg.csv}{9}{0}{\iCIFAR with \Recallvarreg}{\iCIFAR}{fig:icifarSeqRecallVarReg}
	\caption{The results on the \iCIFAR dataset with our four methods.}
	\label{fig:seq_eval_icifar}
	\vspace{0.2cm}
\end{figure*}

\begin{table*}
	\centering 
	\caption{Comparison of \OURAPPROACH and other approaches on iCIFAR-100 dataset, shown in Fig. 5 of the paper}
	\label{tab_comp_cifar100}
	\begin{tabular}{|l|c|c|c|c|c|c|c|c|c|c|}
		\hline
		Approach & Seq $0$ & Seq $1$ & Seq $2$ & Seq $3$ & Seq $4$ & Seq $5$ & Seq $6$ & Seq $7$ & Seq $8$ & Seq $9$ \\ \hline  
		LwF & $95.00$ & $62.05$ & $49.00$ & $42.80$ & $40.04$ & $36.95$ & $33.87$ & $31.36$ & $29.87$ & $27.93$ \\ \hline 
		EWC & $95.00$ & $47.05$ & $31.47$ & $23.58$ & $18.88$ & $15.53$ & $13.30$ & $11.59$ & $10.60$ & $9.67$ \\ \hline 
		SI & $95.00$ & $47.00$ & $31.70$ & $23.48$ & $18.80$ & $15.13$ & $13.24$ & $11.61$ & $10.55$ & $9.62$ \\ \hline 
		AR1 & $76.20$ & $66.35$ & $61.46$ & $56.45$ & $53.20$ & $49.90$ & $46.11$ & $44.40$ & $43.50$ & $42.39$ \\ \hline
		\OURAPPROACH & $95.90$ & $85.75$ & $79.03$ & $75.77$ & $73.26$ & $69.93$ & $68.3$ & $66.12$ & $64.68$ & $\bm{62.04}$ \\ \hline 
		\Recallvar & $94.90$ & $85.15$ & $78.77$ & $75.25$ & $72.80$ & $69.50$ & $68.04$ & $65.89$ & $64.38$ & $61.76$ \\ \hline 
		\Recallreg & $95.20$ & $85.50$ & $78.87$ & $74.27$ & $69.72$ & $64.78$ & $63.3$ & $61.33$ & $59.56$ & $57.30$ \\ \hline 
		\Recallvarreg & $94.80$ & $85.90$ & $78.73$ & $74.25$ & $69.68$ & $64.55$ & $63.17$ & $61.26$ & $59.64$ & $57.39$ \\ \hline \hline
		A-GEM & $95.00$ & $71.85$ & $56.50$ & $54.90$ & $44.90$ & $44.40$ & $37.93$ & $36.94$ & $34.85$ & $25.83$ \\ \hline 
		iCarl & $91.00$ & $83.15$ & $77.07$ & $62.25$ & $44.50$ & $35.40$ & $30.90$ & $21.77$ & $24.61$ & $49.68$ \\ \hline 
	\end{tabular}
		
\end{table*}

\begin{figure*}
	\plotSequenceFig{Plots/Sequence_evaluation/hows_Recall.csv}{4}{0}{\DLRDataset with \OURAPPROACH}{\DLRDataset}{fig:howsSeqRecall}
	\hfill
	\plotSequenceFig{Plots/Sequence_evaluation/hows_Recall_reg.csv}{4}{0}{\DLRDataset with \Recallreg}{\DLRDataset}{fig:howsSeqRecallReg}
	
	\plotSequenceFig{Plots/Sequence_evaluation/hows_Recall_var.csv}{4}{0}{\DLRDataset with \Recallvar}{\DLRDataset}{fig:howsSeqRecallVar}
	\hfill
	\plotSequenceFig{Plots/Sequence_evaluation/hows_Recall_var_reg.csv}{4}{0}{\DLRDataset with \Recallvarreg}{\DLRDataset}{fig:howsSeqRecallVarReg}
	\caption{The results on the \DLRDataset dataset with our four methods.}
	\label{fig:seq_eval_hows}
\end{figure*}

\begin{table}
	\centering 
	\caption{Comparison of \OURAPPROACH and other approaches on \DLRDataset dataset, shown in Fig. 5 of the paper.}
	\label{tab_comp_hows}
	\resizebox{\columnwidth}{!}{		
		\begin{tabular}{|l|c|c|c|c|c|}
			\hline
			Approach & Seq $0$ & Seq $1$ & Seq $2$ & Seq $3$ & Seq $4$ \\ \hline 
			LwF & $96.91$ & $51.91$ & $34.49$ & $27.91$ & $25.13$ \\ \hline 
			EWC & $96.90$ & $57.72$ & $34.91$ & $24.90$ & $17.99$ \\ \hline 
			SI & $96.91$ & $48.83$ & $31.42$ & $23.12$ & $16.25$ \\ \hline 
			AR1 & $28.54$ & $22.33$ & $14.97$ & $11.09$ & $8.59$ \\ \hline
			\OURAPPROACH & $95.71$ & $87.62$ & $78.58$ & $69.73$ & $58.21$ \\ \hline
			\Recallvar & $97.07$ & $86.15$ & $76.06$ & $68.10$ & $\bm{58.34}$ \\ \hline 
			\Recallreg & $97.05$ & $84.01$ & $73.92$ & $66.85$ & $58.13$  \\ \hline 
			\Recallvarreg & $96.73$ & $84.57$ & $74.12$ & $66.94$ & $58.31$ \\ \hline \hline 
			A-GEM & $96.91$ & $87.00$ & $82.60$ & $73.78$ & $61.18$ \\ \hline
			iCarl & $95.71$ & $92.98$ & $89.09$ & $82.66$ & $\bm{73.41}$ \\ \hline 
		\end{tabular}	
	}
\end{table}

\begin{figure*}
	\plotSequenceFig{Plots/Sequence_evaluation/hows-long_Recall.csv}{11}{0}{\DLRDataset with \OURAPPROACH}{\DLRDataset}{fig:howsLongSeqRecall}
	\hfill
	\plotSequenceFig{Plots/Sequence_evaluation/hows-long_Recall_reg.csv}{11}{0}{\DLRDataset with \Recallreg}{\DLRDataset}{fig:howsLongSeqRecallReg}
	
	\plotSequenceFig{Plots/Sequence_evaluation/hows-long_Recall_var.csv}{11}{0}{\DLRDataset with \Recallvar}{\DLRDataset}{fig:howsLongSeqRecallVar}
	\hfill
	\plotSequenceFig{Plots/Sequence_evaluation/hows-long_Recall_var_reg.csv}{11}{0}{\DLRDataset with \Recallvarreg}{\DLRDataset}{fig:howsLongSeqRecallVarReg}
	\caption{The results on the \DLRDataset long dataset with our four methods.}
	\label{fig:seq_eval_hows_long}
\end{figure*}

\begin{table*}
	\centering 
	\caption{Comparison of \OURAPPROACH and other approaches on \DLRDataset long dataset, shown in Fig. 5 of the paper}
	\label{tab_comp_hows_long}
	\resizebox{\textwidth}{!}{	
	\begin{tabular}{|l|c|c|c|c|c|c|c|c|c|c|c|c|}
		\hline
		Approach & Seq $0$ & Seq $1$ & Seq $2$ & Seq $3$ & Seq $4$ & Seq $5$ & Seq $6$ & Seq $7$ & Seq $8$ & Seq $9$ & Seq $10$ & Seq $11$ \\ \hline 	
		LwF & $97.82$ & $49.90$ & $33.41$ & $25.04$ & $20.11$ & $16.98$ & $14.30$ & $12.29$ & $10.85$ & $11.38$ & $11.82$ & $10.29$ \\ \hline 
		EWC & $97.82$ & $49.61$ & $33.33$ & $24.84$ & $19.98$ & $16.62$ & $13.00$ & $12.48$ & $10.63$ & $9.62$ & $9.03$ & $7.01$ \\ \hline 
		SI & $97.82$ & $49.68$ & $33.33$ & $24.87$ & $19.98$ & $16.62$ & $13.10$ & $12.44$ & $10.63$ & $9.40$ & $9.10$ & $6.96$ \\ \hline 
		AR1 & $38.35$ & $20.12$ & $21.92$ & $16.93$ & $16.09$ & $13.39$ & $11.80$ & $12.70$ & $11.05$ & $10.02$ & $9.24$ & $8.40$ \\ \hline
		\OURAPPROACH & $96.95$ & $69.42$ & $54.26$ & $58.51$ & $57.64$ & $58.14$ & $50.89$ & $50.97$ & $46.8$ & $43.11$ & $40.59$ & $37.36$ \\ \hline
		\Recallvar & $96.61$ & $67.41$ & $54.51$ & $58.10$ & $57.96$ & $58.14$ & $50.16$ & $50.74$ & $46.55$ & $42.90$ & $40.30$ & $37.10$ \\ \hline 
		\Recallreg & $98.30$ & $71.40$ & $73.36$ & $72.71$ & $66.79$ & $66.47$ & $56.97$ & $58.23$ & $53.33$ & $49.06$ & $46.78$ & $42.90$ \\ \hline
		\Recallvarreg & $98.04$ & $81.59$ & $79.10$ & $78.77$ & $69.49$ & $68.47$ & $63.25$ & $62.30$ & $57.09$ & $51.70$ & $48.68$ & $\bm{44.62}$  \\ \hline \hline
		A-GEM & $98.72$ & $91.50$ & $33.30$ & $84.29$ & $75.76$ & $82.22$ & $67.02$ & $63.80$ & $73.36$ & $65.08$ & $43.22$ & $\bm{59.46}$ \\ \hline  
		iCarl & $97.48$ & $95.42$ & $95.12$ & $91.87$ & $81.80$ & $65.45$ & $50.42$ & $47.32$ & $45.65$ & $50.38$ & $64.25$ & $56.66$ \\ \hline 	
	\end{tabular}
	}
\end{table*}

\section{\DLRDataset}\label{dataset}

The \DLRDataset dataset has two different versions: the standard one with five sequences and a long version with twelve sequences.
Both versions contain the same images and categories.
In \cref{fig:howsStandard}, the standard version is shown.
Here in each sequence five new categories are introduced to the method. 
These categories are only available during this one sequence and are not stored for later usage.
Furthermore, we use different instances of the same category for testing and training, requiring the method to understand the semantic meaning of a category, rather than just recognizing a certain instance of a category.

In the long version of the \DLRDataset dataset, we start with three categories in the first sequence and then go down to two categories per sequence for the remaining sequences.
This is depicted in \cref{fig:howsLong}.
The long version contains twelve sequences in total, which is the highest amount of sequences of all datasets for continual learning, as far as the authors of this work are aware of. 
Making the learning even more challenging as it firstly contains more training steps, which means more risk to suffer from catastrophic forgetting. 
On top of that, the comparison in each sequence between the different categories is limited, as each sequence now only contains two different categories.
\OURAPPROACH tries to combat this with our so-called recall label $r$.

In contrast to existing datasets like \coreFifty, the network can not learn the instances by heart or even rely on the scene's background to get information about the object.
This is especially difficult as the \DLRDataset even contains categories like fork and spoon, which are similar by design. 
If one compares the images in sequence eight and eleven of the spoon and fork, one will see that this is particularly challenging.
The design decision behind this was to lift continual learning to the next level, overcoming the datasets, where a network could learn based on the environment or a specific instance the matching category, as mentioned in the paper.

Finally, synthetic data makes it easier to randomize the environments, which removes the previous mentioned factor in the evaluation.
This is achieved by using over 1000 different materials, which are randomly assigned to the room's floor and walls.
These materials also have randomized physical properties, meaning that the specularity and other factors of the materials are randomly changed even to further increase the randomness of the environment.

For each RGB we also provide a depth, normal, and segmentation map. 
Please find an example in \cref{fig:myDatasetApple}.
A complete list of the different categories used in \DLRDataset is shown in \cref{tab_hows_cat}.

\input{tikz_figures/hows_standard}
\input{tikz_figures/hows_long}

\begin{table}
	\centering 
	\caption{Overview of all categories of our \OURAPPROACH dataset}
	\label{tab_hows_cat}
	\resizebox{\columnwidth}{!}{		
		\begin{tabular}{|l|c|c|}
			\hline
			\textbf{Category} & \textbf{\DLRDataset} & \textbf{\DLRDataset long} \\ \hline 
			Apple & seq. 0 & seq. 0 \\ \hline 
			Bag & seq. 0 & seq. 0 \\ \hline
			Ball & seq. 0 & seq. 0 \\ \hline
			Banana & seq. 0 & seq. 1 \\ \hline
			Bowl & seq. 0 & seq. 1 \\ \hline
			Bread & seq. 1 & seq. 2 \\ \hline
			Camera & seq. 1 & seq. 2 \\ \hline
			Can & seq. 1 & seq. 3 \\ \hline
			Cap & seq. 1 & seq. 3 \\ \hline
			Computer keyboard & seq. 1 & seq. 4 \\ \hline
			Egg & seq. 2 & seq. 4 \\ \hline
			Fork & seq. 2 & seq. 5 \\ \hline
			Glass bottle & seq. 2 & seq. 5 \\ \hline
			Glasses & seq. 2 & seq. 6 \\ \hline
			Headset & seq. 2 & seq. 6 \\ \hline
			Knife & seq. 3 & seq. 7 \\ \hline
			Milk canister & seq. 3 & seq. 7 \\ \hline
			Mobile phone & seq. 3 & seq. 8 \\ \hline
			Mug & seq. 3 & seq. 8 \\ \hline
			Pan & seq. 3 & seq. 9 \\ \hline
			Pear & seq. 4 & seq. 9 \\ \hline
			Pen & seq. 4 & seq. 10 \\ \hline
			Scissors & seq. 4 & seq. 10 \\ \hline
			Spoon & seq. 4 & seq. 11 \\ \hline
			Teddy bear & seq. 4 & seq. 11 \\ \hline
		\end{tabular}	
	}
\end{table}

\section{Additional ablation studies}\label{AddAblationStudies}

\subsection{Number of layers}

In \cref{tab_nrlayers} the resulting accuracies on \coreFifty and \DLRDataset dataset are shown, when different amounts of fully-connected-layers are used.
It is shown that the best results for both datasets are achieved when one fully-connected layer is used, which is our default option of \OURAPPROACH, as more layers do not further improve the results.
We speculate the reason for this is that the features from the ResNet50 are already separated well enough that further processing only hurts the overall performance. 

\begin{table}
	\centering
	\caption{Results of \OURAPPROACH on different datasets when using different amounts of fully-connected layers. The default option is marked with a star.}
	\label{tab_nrlayers}
		\begin{tabular}{| c | c | c |} \hline
			Layer amount & \coreFifty & \DLRDataset \\ \hline
			1* & $\bm{72.02}$ & $\bm{58.31}$ \\ \hline
			2 & $71.65$ & $52.11$ \\ \hline
			3 & $70.32$ & $52.21$ \\ \hline
			4 & $69.95$ & $53.56$ \\ \hline
		\end{tabular}
\end{table}

\section{Hyperparameter and source code}\label{sourceCode}

The exact hyperparameter used to achieve all the reported results of the different \OURAPPROACH versions and the implementation of our approach are available in our GitHub repository: \url{https://github.com/DLR-RM/RECALL}.
The same is true for our \DLRDataset dataset. Which can also be downloaded here: \url{https://zenodo.org/record/7054171}.


%% file: chapters/tikz_commands.tex
\usepackage{tikz}
\usepackage{ifthen}
\usetikzlibrary{decorations.pathreplacing, calc, backgrounds}
\usetikzlibrary{shapes, shadows.blur, arrows.meta, positioning}
\usepackage{pdfrender}
\usepackage{tikzsymbols}
\usepackage{pgfplots}
\usepackage{pgfplotstable}
\pgfplotsset{compat=1.11}

\makeatletter
\newcommand{\gettikzxy}[3]{%
  \tikz@scan@one@point\pgfutil@firstofone#1\relax
  \edef#2{\the\pgf@x}%
  \edef#3{\the\pgf@y}%
}
\makeatother

\usepackage{multirow}
\usepackage{makecell}

\tikzstyle{generalStyle}=[blur shadow={shadow blur steps=7}]

\definecolor{baseClasses}{RGB}{209, 67, 67}
\definecolor{sequenceClasses}{RGB}{1, 102, 153}
\definecolor{hellblue}{RGB}{98,140,178}

\definecolor{secLine}{rgb}{0.050 0.850 1.000} 
\definecolor{thirdLine}{rgb}{0.050 0.380 1.000} 
\definecolor{fourLine}{rgb}{0.500 0.300 0.700} 
\definecolor{fiveLine}{rgb}{1.000 0.050 0.690} 
\definecolor{sixLine}{rgb}{1.000 0.120 0.050} 
\definecolor{sevenLine}{rgb}{1.000 0.610 0.000} 

\newcommand{\drawPlane}[6]{
	\set{\xTempInsideDrawPlane}{#1};
	\set{\yTempInsideDrawPlane}{#2};
	\set{\zTempInsideDrawPlane}{-#3};
	\set{\widthTempInsideDrawPlane}{#4};
	\set{\heightTempInsideDrawPlane}{#5};
	\draw [#6] (\xTempInsideDrawPlane, \yTempInsideDrawPlane, \zTempInsideDrawPlane) -- (\xTempInsideDrawPlane, \yTempInsideDrawPlane+\heightTempInsideDrawPlane, \zTempInsideDrawPlane) -- (\xTempInsideDrawPlane+\widthTempInsideDrawPlane, \yTempInsideDrawPlane+\heightTempInsideDrawPlane, \zTempInsideDrawPlane) -- (\xTempInsideDrawPlane+\widthTempInsideDrawPlane, \yTempInsideDrawPlane, \zTempInsideDrawPlane) -- cycle;
}

\newcommand{\drawIndividualFrustum}[7]{
	\set{\xTempFilledBlock}{#1};
	\set{\yTempFilledBlock}{#2};
	\set{\zTempFilledBlock}{-#3};
	
	\set{\leftHeightTempFilledBlock}{#4};
	\set{\rightHeightTempFilledBlock}{#5};
	\set{\widthTempFilledBlock}{#6};

	\pgfmathparse{\leftHeightTempFilledBlock>\rightHeightTempFilledBlock};
	\set{\isLeftBigger}{\pgfmathresult};
	\ifthenelse{\isLeftBigger=1}{
		\set{\leftAdd}{0};
		\set{\rightAdd}{(\leftHeightTempFilledBlock - \rightHeightTempFilledBlock) * 0.5};
	}{
		\set{\leftAdd}{(\rightHeightTempFilledBlock - \leftHeightTempFilledBlock) * 0.5};
		\set{\rightAdd}{0};
	};
	
	\draw [#7] (\xTempFilledBlock, \yTempFilledBlock+\leftAdd, \zTempFilledBlock) -- 
	(\xTempFilledBlock+\widthTempFilledBlock, \yTempFilledBlock + \rightAdd, \zTempFilledBlock) -- 
	(\xTempFilledBlock+\widthTempFilledBlock, \yTempFilledBlock+\rightHeightTempFilledBlock+\rightAdd, \zTempFilledBlock) -- 
	(\xTempFilledBlock, \yTempFilledBlock+\leftHeightTempFilledBlock+\leftAdd, \zTempFilledBlock) -- cycle;	
}

\newcommand{\drawLegendEntry}[6]{
	\pgfmathsetmacro{\posX}{#1};
	\pgfmathsetmacro{\posY}{#2};
	\pgfmathsetmacro{\width}{#3};
	\pgfmathsetmacro{\useit}{#6};
	\pgfmathsetmacro{\posXX}{\posX + \width};
	\node at (\posX,\posY) (start) {};
	\node[anchor=west] at (\posXX,\posY) (end) {#4}; 
	
	\node at (\posX+0.2,\posY-0.17) (startrect) {};
	\node at (\posXX,\posY+0.17) (endrect)  {};
	\begin{scope}
		\ifthenelse{\useit = 1}{
			\draw[#5, line width=3.5mm, draw opacity=0.25] (start) -- (end);
			
			\draw[#5, line width=0.7mm] (start) -- (end);
		}{
			\draw[#5, line width=0.8mm] (start) -- (end);
		};
	\end{scope}
	
}

\tikzstyle{imgStyle}=[blur shadow={shadow blur steps=7, shadow xshift=1pt, shadow yshift=-1pt}]
\tikzset{
	boxFrame/.style={ minimum width=1.5 cm,%
		minimum height=1.5 cm,%
		align=center}
}

\definecolor{aronecolor}{RGB}{209, 67, 67}
\definecolor{lwfcolor}{RGB}{246, 146, 30}
\definecolor{ewccolor}{RGB}{110, 70, 114}
\definecolor{sicolor}{RGB}{100, 158, 10}
\definecolor{iCarlcolor}{RGB}{1, 102, 153}
\definecolor{aGemcolor}{RGB}{3, 175, 175}

\tikzstyle{ourApproachStyle} = [black, thick, mark=star]
\tikzstyle{aronestyle} = [aronecolor, thick, dashed, mark options={scale=0.75, solid}, mark=diamond*]   
\tikzstyle{aronestylereproduced} = [aronecolor, thick, mark options={scale=0.75}, mark=diamond*] 
\tikzstyle{lwfStyle} = [lwfcolor, thick, mark=|]
\tikzstyle{ewcStyle} = [ewccolor, thick, mark=|]
\tikzstyle{siStyle} = [sicolor, thick, mark=|]
\tikzstyle{iCarlStyle} = [iCarlcolor, mark options={scale=0.75}, only marks, mark=*]
\tikzstyle{aGemStyle} = [aGemcolor, mark options={scale=0.75}, only marks, mark=square]

\newcommand{\plotGraph}[4]{
\ifthenelse{#4 = 1}{
	\addplot [aronestylereproduced] table[x=Sequence, y=AR1, col sep=comma] {#1};  
}{
	\addlegendimage{aronestylereproduced}
}
\ifthenelse{#2 = 1}{\addlegendentry{AR1} }{}												
\ifthenelse{#4 = 1}{
	\addplot [lwfStyle] table[x=Sequence, y=LwF, col sep=comma] {#1};
}{
	\addlegendimage{lwfStyle}	
}
\ifthenelse{#2 = 1}{\addlegendentry{LwF} }{}
\ifthenelse{#4 = 1}{
	\addplot [ewcStyle] table[x=Sequence, y=EWC, col sep=comma] {#1};
}{
	\addlegendimage{ewcStyle}	
}
\ifthenelse{#2 = 1}{\addlegendentry{EWC} }{}
\ifthenelse{#4 = 1}{
	\addplot [siStyle] table[yellow, x=Sequence, y=SI, col sep=comma, mark=|] {#1};
}{
	\addlegendimage{siStyle}	
}
\ifthenelse{#2 = 1}{\addlegendentry{SI} }{}
\ifthenelse{#3 = 1}{
	\ifthenelse{#4 = 1}{
		\addplot [iCarlStyle] table[x=Sequence, y=iCarl, col sep=comma] {#1};
	}{
		\addlegendimage{iCarlStyle}
	}
	\ifthenelse{#2 = 1}{\addlegendentry{\iCaRL} }{}
	\ifthenelse{#4 = 1}{
		\addplot [aGemStyle] table[x=Sequence, y=A-GEM, col sep=comma] {#1};
	}{
		\addlegendimage{aGemStyle}	
	}
	\ifthenelse{#2 = 1}{\addlegendentry{A-GEM} }{}
}{}
\ifthenelse{#4 = 1}{
	\addplot [ourApproachStyle] table[x=Sequence, y=Our Approach, col sep=comma] {#1};
}{
	\addlegendimage{ourApproachStyle}
}
\ifthenelse{#2 = 1}{\addlegendentry{\OURAPPROACH}}{}

}

\newcommand{\plotSequence}[4]{
  \begin{tikzpicture}[mark=*]
	\begin{axis}[
	colormap={colorList}{
     rgb(1)=(0.702, 0.0, 0.0);
     rgb(2)=(0.235, 0.706, 0.294);
     rgb(3)=(0.263, 0.388, 0.847);
     rgb(4)=(0.961, 0.51, 0.192);
     rgb(5)=(0.259, 0.831, 0.957);
     rgb(6)=(0.941, 0.196, 0.902);
     rgb(7)=(0.275, 0.6, 0.565);
     rgb(8)=(0.043, 0, 0);
     rgb(9)=(0.0, 0.0, 0.459);
     rgb(10)=(0.663, 0.663, 0.663);
     rgb(11)=(0.667, 1, 0.765);
     rgb(12)=(0.0, 0.0, 0.0);
    },
	cycle list={[indices of colormap={0,...,12} of colorList]},
	title={Results on #4 for each validation sequence},
	xtick=data, legend columns=5,
	title style={at={(-0.3,0.5)},anchor=north, rotate=90},
	legend style={column sep=5pt, at={(0.45, 1.1)}, anchor=south, nodes={scale=0.7, transform shape}},
	ymin=#3,
	ymax=100,
	ylabel={Accuracy in \%}]
	\foreach \i in {0,...,#2}{
		\edef\addLegendCurrent{\noexpand\addlegendentry{Seq {\i}}};
		\addplot table[ col sep=comma, x expr=\coordindex, y=\i] {#1};
		\addLegendCurrent;
	}
	\end{axis}
  \end{tikzpicture}
}


%% file: chapters/commands.tex
\usepackage{xargs} 
\usepackage{caption}
\usepackage{algorithm} 
\usepackage[noend]{algpseudocode}
\usepackage{bm} 
\usepackage{mathbbol}

\usepackage{varwidth}
\usepackage{subcaption}

\newcommand{\set}[2]{\pgfmathsetmacro{#1}{#2}}

\tikzset{
    max width/.style args={#1}{
        execute at begin node={\begin{varwidth}{#1}},
        execute at end node={\end{varwidth}}
    }
}

\makeatletter
\DeclareRobustCommand\onedot{\futurelet\@let@token\@onedot}
\def\@onedot{\ifx\@let@token.\else.\null\fi\xspace}

\def\eg{\emph{e.g}\onedot}

\def\etal{\emph{et al}\onedot}
\makeatother

\newcommand{\DLRDataset}{HOWS-CL-25\xspace}
\newcommand{\coreFifty}{CORe50\xspace}

\newcommand{\iCIFAR}{iCIFAR-100\xspace}

\newcommand{\iCaRL}{iCaRL\xspace}

\newcommand{\reconstructionLabelWord}{recall label\xspace} 
\newcommand{\reconstructionLabelWordPlural}{recall labels\xspace}
\newcommand{\reconstructionLossName}{full regression loss\xspace} 
\newcommand{\OURAPPROACH}{RECALL\xspace}
\newcommand{\parenth}[1]{\left(#1\right)}
\newcommand{\brackets}[1]{\left[#1\right]}

\newcommand{\sequence}{s}
\newcommand{\prevSequence}{\hat{\sequence}}
\newcommand{\sequences}{S}
\newcommand{\class}{c}
\newcommand{\classes}{C}
\newcommand{\networkWithoutX}{f\xspace}
\newcommand{\network}{\networkWithoutX\left(x\right)}
\newcommand{\networkoutput}{o}
\newcommand{\image}{x}

\newcommand{\images}{x}
\newcommand{\olabel}{y}

\newcommand{\olabels}{y}
\newcommand{\reconstructionlabel}{r}
\newcommand{\weights}{\theta}

\newcommand{\trainingsexample}{i}

\newcommand{\classknown}{\makeknown{\classes}_{\sequence}}
\newcommand{\trainingsexampleAndSequence}{_{\trainingsexample,\sequence}}

\newcommand{\forTrainingExampleAndClass}{_{\trainingsexample}[\class]}
\newcommand{\networkoutputClipped}{\bar{\networkoutput}}
\newcommand{\layerName}[2]{\text{fc}\!\brackets{#1, #2}}
\newcommand{\layerNameBroken}[4]{$\text{fc}\!\brackets{#1,\!#2}$ \\ #4$\,\!\parenth{#3}$}

\newcommand{\binaryIndicator}{\mathbb{1}}
\newcommand{\predictionPropability}{p}
\newcommand{\loss}{L} 

\newcommand{\makeknown}[1]{{#1}^{\mkern 1mu \text{prev}}}
\newcommand{\makerecall}[1]{{#1}^{\mkern 1mu \text{recall}}}
\newcommand{\makenew}[1]{{#1}^{\mkern 1mu \text{curr}}}
\newcommand{\makeall}[1]{{#1}^{\mkern 1mu \text{all}}}
\newcommand{\lossonoldweights}{\makerecall{\loss}}   
\newcommand{\lossonnewweights}{\makenew{\loss}}	 
\newcommand{\lossonallweights}{\makeall{\loss}}	 
\newcommand{\lossonoldweightsvariance}{\makerecall{\bar{\loss}}}
\newcommand{\reconstructionLoss}{\tilde{\loss}}
\newcommand{\reconstructionLossAllWeights}{\makeall{\reconstructionLoss}}
\newcommand{\reconstructionLossNewWeights}{\makenew{\reconstructionLoss}}

\newcommand{\varianz}{\sigma} 
\newcommand{\mean}{\mu}

\newcommand{\moveUpInImage}{\vspace{-0.15cm}}

%% file: chapters/abstract.tex
Convolutional neural networks show remarkable results in classification but struggle with learning new things on the fly.
We present a novel rehearsal-free approach, where a deep neural network is continually learning new unseen object categories without saving any data of prior sequences. 
Our approach is called \OURAPPROACH, as the network recalls categories by calculating logits for old categories before training new ones.
These are then used during training to avoid changing the old categories.
For each new sequence, a new head is added to accommodate the new categories. 
To mitigate forgetting, we present a regularization strategy where we replace the classification with a regression.
Moreover, for the known categories, we propose a Mahalanobis loss that includes the variances to account for the changing densities between known and unknown categories.
Finally, we present a novel dataset for continual learning (\DLRDataset), especially suited for object recognition on a mobile robot, including 150,795 synthetic images of 25 household object categories.
Our approach \OURAPPROACH outperforms the current state of the art on \coreFifty and \iCIFAR and reaches the best performance on \DLRDataset.

%% file: chapters/introduction.tex
Humans are remarkable in extracting new knowledge about unknown things continuously throughout their lifetime. 
Thus, lifelong learning is a crucial capability in our daily life.
Deep neural networks have shown excellent results on many problems, from recognition to reconstruction tasks in computer vision and more \cite{He2016, Sundermeyer2018, Brachmann2018, Graves2014}.
Typically, these algorithms apply batch-wise training to large datasets, \eg, ImageNet by Deng \etal \cite{Deng2009}, and then need many iterations over the whole dataset to obtain satisfactory performance.  
In contrast to humans, neural networks rely on a static dataset, which has to be fixed before the training starts.
\input{tikz_figures/first_page.tex}
For new categories that have not been part of the original training set, the network has to be trained again, while the original training set has to be kept in memory (rehearsal) to prevent forgetting.
It is not practical to keep relearning previously seen categories as it is time and memory-consuming.
So in this work, we present a novel approach to do \textbf{Re}hearsal-free \textbf{C}ontinu\textbf{al} \textbf{L}earning called \textbf{\OURAPPROACH}, which can be used to enable a network to learn new categories on-demand without the need of retraining the whole network.
Therefore, the experiments are designed for a class-incremental learning (Class-IL) \cite{VandeVen2019}, schematized in \cref{fig:introPicture}, where each sequence contains a collection of images of a non-reoccurring fixed amount of categories. This means that a category is only present in one sequence.
In order to get closer to a lifelong learning approach, we propose a rehearsal-free approach, which does not require storing any visual history like most other solutions in literature \cite{Rebuffi2017, LopezPaz2017}.
The main contributions of this work are:
\begin{itemize}
	\item Proposal of \OURAPPROACH: A method to mitigate forgetting in a rehearsal-free continual learning environment.
	\item Analysis of the logit output distribution discrepancies of \OURAPPROACH, which are caused by learning over different sequences. Evaluation and proposal of different solutions to cope with these discrepancies.
	\item Introduction of a novel dataset (\DLRDataset) with household objects and two suggested sequence structures in a continual learning manner.	
\end{itemize}

%% file: tikz_figures/first_page.tex
\begin{figure}[t]
\centering
\resizebox{0.9757\columnwidth}{!}{
\begin{tikzpicture}

\set{\xstart}{0}
\set{\ystart}{0}
\set{\lineLength}{7}
\set{\xSeqFirstPos}{1.25}
\set{\imageSize}{1.4}
\set{\xSeqSecondPos}{\lineLength - \xSeqFirstPos}
\set{\arrowSpacing}{0.02};
\set{\classAmount}{5};
\set{\secondClassAmount}{3};

\node[inner sep=0pt] (lineStart) at (\xstart, \ystart) {};
\node[inner sep=0pt] (lineEnd) at (\xstart + \lineLength, \ystart) {};

\draw[->, thick] (lineStart) -> (lineEnd);
\node [right= 0.05cm of lineEnd] {time};

\node [circle, draw, inner sep=1.5pt, fill] at (\xstart + \xSeqFirstPos, \ystart) (firstCircle) {};
\node [anchor=north, below= 0.1cm of firstCircle] (seqzerotext) {Seq $0$};
\node [circle, draw, inner sep=1.5pt, fill] at (\xstart + \xSeqSecondPos, \ystart) (secondCircle) {};
\node [anchor=north, below= 0.1cm of secondCircle] (seqonetext) {Seq $1$};
\node [anchor=north, right= 0.1cm of seqonetext] {$\hdots$};

\node [inner sep=0pt, above= 0.15cm of firstCircle, anchor=south] (firstImgFirst)  {\includegraphics[width=\imageSize cm]{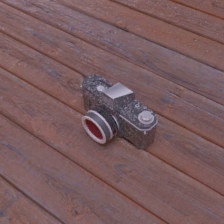}};
\node [imgStyle, inner sep=0pt, above right= 0.1cm and 0.05cm of firstImgFirst, anchor=south] (firstImgSecond) {\includegraphics[width=\imageSize cm]{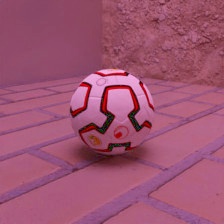}};
\node [imgStyle, inner sep=0pt, above left= 0.1cm and 0.05cm of firstImgFirst, anchor=south] (firstImgThird) {\includegraphics[width=\imageSize cm]{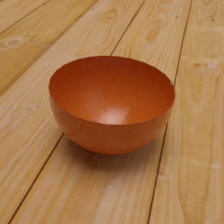}};
\node [imgStyle, inner sep=0pt, left= 0.1cm of firstImgFirst, anchor=east] (firstImgFourth) {\includegraphics[width=\imageSize cm]{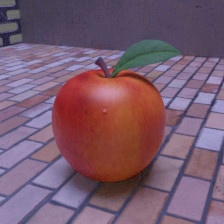}};
\node [imgStyle, inner sep=0pt, right= 0.1cm of firstImgFirst, anchor=west] (firstImgFivth) {\includegraphics[width=\imageSize cm]{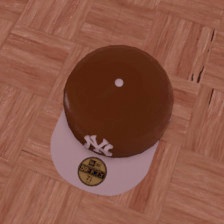}};

\node [imgStyle, inner sep=0pt, above= 0.1cm of secondCircle, anchor=south] (secondImgFirst) {\includegraphics[width=\imageSize cm]{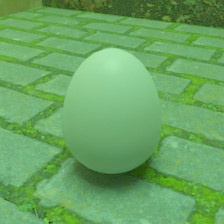}};
\node [imgStyle, inner sep=0pt, above right= 0.1cm and 0.05cm of secondImgFirst, anchor=south] (secondImgSecond) {\includegraphics[width=\imageSize cm]{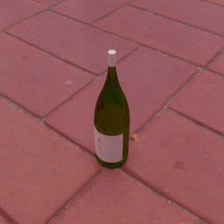}};
\node [imgStyle, inner sep=0pt, above left= 0.1cm and 0.05cm of secondImgFirst, anchor=south] (secondImgThird) {\includegraphics[width=\imageSize cm]{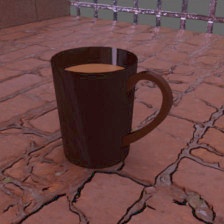}};

\set{\distBetween}{0.25};
\set{\orgBoxHeight}{3};
\set{\filterBoxFactor}{0.8};
\set{\boxHeight}{(\orgBoxHeight-\distBetween) / 2  / \filterBoxFactor};
\set{\smallBoxHeight}{\boxHeight};
\set{\boxSpacing}{0.85};
\set{\filterAmountFirst}{5};
\set{\filterAmount}{2};
\set{\filterWidth}{0.3};
\set{\extractionWidth}{0.75};
\set{\yLeftBoxValue}{\ystart - 0.66};
\set{\spacingResNet}{0.2};
\set{\filterHeight}{(\orgBoxHeight*\filterBoxFactor) / 2-\distBetween*0.5};
\set{\spacing}{0.3};
\set{\spacingHalf}{\spacing*0.5};
\set{\boxWidth}{\spacing*4.2+\filterAmount*\filterWidth+\extractionWidth+\imageSize};
\set{\xBoxStart}{\xstart+\xSeqFirstPos-\boxWidth/2};
\set{\filterStart}{\xBoxStart+\spacing-\spacingHalf*0.5+\imageSize+\extractionWidth};

\newcommand{\nrNeurons}{{2048, \classAmount}};
\set{\lowerYPos}{\yLeftBoxValue - \boxHeight - \boxSpacing + (\boxHeight - \filterHeight) / 2};

\drawPlane{\xstart+\xSeqFirstPos-\boxWidth/2}{\yLeftBoxValue - \boxHeight - \boxSpacing}{0}{\boxWidth}{\boxHeight}{generalStyle, fill=white, draw=black, rounded corners=1mm};
\foreach \layerNr in {1, ..., \filterAmount}{
	\set{\xPos}{\layerNr * \spacing + (\layerNr - 1) * \filterWidth + \filterStart};
	\ifthenelse{\layerNr = 1}{
		\set{\lastXPos}{\xPos - \spacing * 0.6125};	
	}{
		\set{\lastXPos}{\xPos - \spacing};	
	}
	\def\fillColor{lightgray};
	\ifthenelse{\layerNr = 2}{\def\fillColor{baseClasses}}{};
	\drawPlane{\xPos}{\lowerYPos}{0}{\filterWidth}{\filterHeight}{imgStyle, draw=black, rounded corners=0.75mm, fill=\fillColor};
	\draw [->] (\lastXPos + \arrowSpacing, \lowerYPos+\filterHeight*0.5) -- (\xPos - \arrowSpacing, \lowerYPos+\filterHeight*0.5);
	
	\pgfmathparse{\nrNeurons[\layerNr - 1]}
	\set{\nrCurrentNeurons}{int(\pgfmathresult)};	
	\node [anchor=north,yshift=1.5pt] at (\xPos+\filterWidth/2, \lowerYPos) {\tiny (\nrCurrentNeurons)};	
}
\set{\xPos}{\filterStart+\filterWidth+\spacing*2};
\set{\classAmountMinusOne}{\classAmount - 1}
\foreach \currentlayerNr in {1, ..., \classAmountMinusOne}{
	\set{\xPosEnd}{\xPos + \filterWidth};
	\set{\yPos}{\lowerYPos + (\filterHeight / \classAmount) * \currentlayerNr};	
	\draw (\xPos, \yPos) -- (\xPosEnd, \yPos);
	\node [inner sep=0pt, anchor=west, xshift=2.5pt] at (\xPosEnd, \yPos-\filterHeight / \classAmount*0.5) {\tiny $0$};
}

\node [inner sep=0pt, anchor=west, xshift=2.5pt] at (\xPos + \filterWidth, \lowerYPos+\filterHeight-\filterHeight / \classAmount*0.5) {\tiny $1$};

\node [inner sep=0pt] at (\xBoxStart, \yLeftBoxValue - \boxHeight / 2 - \boxSpacing) (boxStart) {};
\node [inner sep=0pt] at (\xBoxStart+\boxWidth/2, \yLeftBoxValue - \boxHeight - \boxSpacing) (boxSouth) {};

\node [imgStyle, inner sep=0pt, right= \spacingHalf cm of boxStart, anchor=west] (firstImgFourthCopied) {\includegraphics[width=\imageSize cm]{pictures/new_classes/class_0.jpg}};

\gettikzxy{(firstImgFourthCopied.east)}{\ax}{\ay}
\set{\axA}{\ax/1cm};
\drawIndividualFrustum{\axA+\spacingHalf}{\yLeftBoxValue - \boxHeight / 2 - \boxSpacing - \imageSize /2}{0}{\imageSize}{\extractionWidth*0.6}{\extractionWidth}{imgStyle, rounded corners=0.3mm,fill=hellblue};

\node [inner sep=0pt, text ragged, align=center, font=\tiny\linespread{0.8}\selectfont] at (\axA+\spacingHalf+\extractionWidth*0.5, \yLeftBoxValue - \boxHeight / 2 - \boxSpacing) {\tiny frozen \\ \tiny ResNet50};

\node [align=center, below= 0.05cm of boxSouth.south, anchor=north] {Train};

\set{\boxHeight}{\orgBoxHeight};
\set{\filterHeight}{\filterBoxFactor * \boxHeight};
\set{\spacing}{0.25};
\set{\spacingHalf}{\spacing*0.5};
\set{\boxWidth}{\spacing*4.2+\filterAmount*\filterWidth+\extractionWidth+\imageSize+\spacingResNet};
\set{\xBoxStart}{\xstart+\xSeqSecondPos-\boxWidth/2};
\set{\filterStart}{\xBoxStart+\spacing-\spacingHalf*0.5+\spacingResNet+\imageSize+\extractionWidth};
\set{\lowerYPos}{\ystart - \boxHeight - \boxSpacing + (\boxHeight - \filterHeight) / 2};

\drawPlane{\xstart+\xSeqSecondPos-\boxWidth/2}{\ystart - \boxHeight - \boxSpacing}{0}{\boxWidth}{\boxHeight}{generalStyle, fill=white, draw=black, rounded corners=1mm};

\newcommand{\nrCategories}{{\secondClassAmount, \classAmount}};
\set{\sumClasses}{\classAmount + \secondClassAmount};
\foreach \layerNr in {1, ..., \filterAmount}{
	\set{\xPos}{\layerNr * \spacing + (\layerNr - 1) * \filterWidth + \filterStart};

	\ifthenelse{\layerNr = 2}{\def\fillColor{baseClasses}}{};
	\ifthenelse{\layerNr = 2}{	
		\foreach \partNr in {0, ..., 1}{
			\pgfmathparse{\nrCategories[\partNr]}
			\set{\nrCurrentNeurons}{\pgfmathresult};			
			\ifthenelse{\partNr = 0}{
				\def\fillColor{sequenceClasses}
				\set{\ySize}{\nrCurrentNeurons / \sumClasses * (\filterHeight - \distBetween)};
				\set{\yOffset}{((\filterHeight *0.5 - \distBetween * 0.5) + \ySize) / 2 - \ySize};
				\set{\currentClass}{\secondClassAmount};
			}{
				\def\fillColor{baseClasses};
				\set{\yOffset}{0.5 * \filterHeight + \distBetween * 0.5};
				\set{\ySize}{0.5 * (\filterHeight - \distBetween)};
				\set{\currentClass}{\classAmount};
			};
			\drawPlane{\xPos}{\lowerYPos+ \yOffset}{0}{\filterWidth}{\ySize}{imgStyle, draw=black, rounded corners=0.75mm, fill=\fillColor};		
			\node [anchor=north,yshift=1.5pt] at (\xPos+\filterWidth/2, \lowerYPos+ \yOffset) {\tiny (\currentClass)};
		}
	}{
		\drawPlane{\xPos}{\lowerYPos+\filterHeight*0.5+\distBetween*0.5}{0}{\filterWidth}{\filterHeight*0.5-\distBetween*0.5}{imgStyle, draw=black, rounded corners=0.75mm, fill=lightgray};
		\drawPlane{\xPos}{\lowerYPos}{0}{\filterWidth}{\filterHeight*0.5-\distBetween*0.5}{imgStyle, draw=black, rounded corners=0.75mm, fill=lightgray};
		
		\node [anchor=north,yshift=1.5pt] at (\xPos+\filterWidth/2, \lowerYPos) {\tiny (2048)};
		\node [anchor=north,yshift=1.5pt] at (\xPos+\filterWidth/2, \lowerYPos+\filterHeight*0.5+\distBetween*0.5) {\tiny (2048)};
	}	
	\set{\yLowerMiddle}{\lowerYPos+(\filterHeight*0.5-\distBetween*0.5)*0.5};
	\set{\yUpperMiddle}{\lowerYPos+\filterHeight-(\filterHeight*0.5-\distBetween*0.5)*0.5};

	\ifthenelse{\layerNr = 1}{
		\set{\lastXPos}{\xPos - \spacing * 0.6125 - \spacingResNet};		
		\set{\yUpperStart}{\lowerYPos+\filterHeight*0.5};	
		\set{\yLowerStart}{\lowerYPos+\filterHeight*0.5};
	}{
		\set{\lastXPos}{\xPos - \spacing};	
		\set{\yUpperStart}{\yUpperMiddle};	
		\set{\yLowerStart}{\yLowerMiddle};
	}

	\draw [->] (\lastXPos + \arrowSpacing, \yLowerStart)  to [out=0,in=180] (\xPos - \arrowSpacing, \yLowerMiddle);
	\draw [->] (\lastXPos + \arrowSpacing, \yUpperStart)  to [out=0,in=180] (\xPos - \arrowSpacing, \yUpperMiddle);

}
\set{\xPos}{\filterStart+\filterWidth+\spacing*2};

\foreach \partNr in {0, ..., 1}{
	\pgfmathparse{\nrCategories[\partNr]}
	\set{\nrCurrentNeurons}{\pgfmathresult};		
	\set{\classAmountMinusOne}{\nrCurrentNeurons - 1}
	\set{\currentFilterHeight}{\nrCurrentNeurons / \sumClasses * \filterHeight};
	\ifthenelse{\partNr = 0}{
		\def\fillColor{sequenceClasses}
		\set{\ySize}{\nrCurrentNeurons / \sumClasses * (\filterHeight - \distBetween)};
		\set{\yOffset}{((\filterHeight *0.5 - \distBetween * 0.5) + \ySize) / 2 - \ySize};
		\node [inner sep=0pt, anchor=west, xshift=2.5pt] at (\xPos + \filterWidth, \lowerYPos+\yOffset + \ySize - \ySize / \nrCurrentNeurons*0.5) {\tiny $1$};
	}{
		\def\fillColor{baseClasses};
		\set{\yOffset}{0.5 * \filterHeight + \distBetween * 0.5};
		\set{\ySize}{0.5 * (\filterHeight - \distBetween)};
		\node [inner sep=0pt, anchor=west, xshift=2.5pt] at (\xPos + \filterWidth, \lowerYPos+ \ySize/2+\yOffset) (recLabel) {\tiny $\reconstructionlabel_{\sequence}$};

	};
	\foreach \currentlayerNr in {1, ..., \classAmountMinusOne}{
		\set{\xPosEnd}{\xPos + \filterWidth};
		
		\set{\yPos}{\lowerYPos + \yOffset + (\ySize / \nrCurrentNeurons) * \currentlayerNr};	
		\draw (\xPos, \yPos) -- (\xPosEnd, \yPos);
		\ifthenelse{\partNr = 0}{
		\node [inner sep=0pt, anchor=west, xshift=2.5pt] at (\xPosEnd, \yPos-\filterHeight / \sumClasses*0.5) {\tiny $0$};}{}
	}
}

\node [inner sep=0pt] at (\xBoxStart, \ystart - \boxHeight / 2 - \boxSpacing) (boxStart) {};
\node [inner sep=0pt] at (\xBoxStart+\boxWidth/2, \ystart - \boxHeight - \boxSpacing) (boxSouth) {};

\node [imgStyle, inner sep=0pt, right= \spacingHalf cm of boxStart, anchor=west] (secondImgFirstCopied) {\includegraphics[width=\imageSize cm]{pictures/new_classes/class_5.jpg}};

\gettikzxy{(secondImgFirstCopied.east)}{\ax}{\ay}
\set{\axA}{\ax/1cm};
\drawIndividualFrustum{\axA+\spacingHalf}{\ystart - \boxHeight / 2 - \boxSpacing - \imageSize /2}{0}{\imageSize}{\extractionWidth*0.6}{\extractionWidth}{imgStyle, rounded corners=0.3mm,fill=hellblue};

\node [inner sep=0pt, text ragged, align=center, font=\tiny\linespread{0.8}\selectfont] at (\axA+\spacingHalf+\extractionWidth*0.5, \ystart - \boxHeight / 2 - \boxSpacing) {\tiny frozen \\ \tiny ResNet50};

\node [align=center, below= 0.05cm of boxSouth.south, anchor=north] {Train};

\draw [->, thick] (firstImgFourth.south) to [out=270,in=90] (firstImgFourthCopied.north); 

\draw [->, thick] (secondImgFirst.west) to [out=180,in=90] (secondImgFirstCopied.north); 
\set{\xJustinPos}{\xstart + \xSeqFirstPos - \spacing*0.7};
\set{\imageSize}{\imageSize*1.4};
\set{\yJustinLower}{\ystart - \smallBoxHeight - \boxSpacing - \spacing*2.9 - \imageSize}
\set{\diffBetweenJustins}{\imageSize*0.8}
\node [inner sep=0pt, anchor=south] (justin) at (\xJustinPos, \yJustinLower) {\includegraphics[width=\imageSize cm]{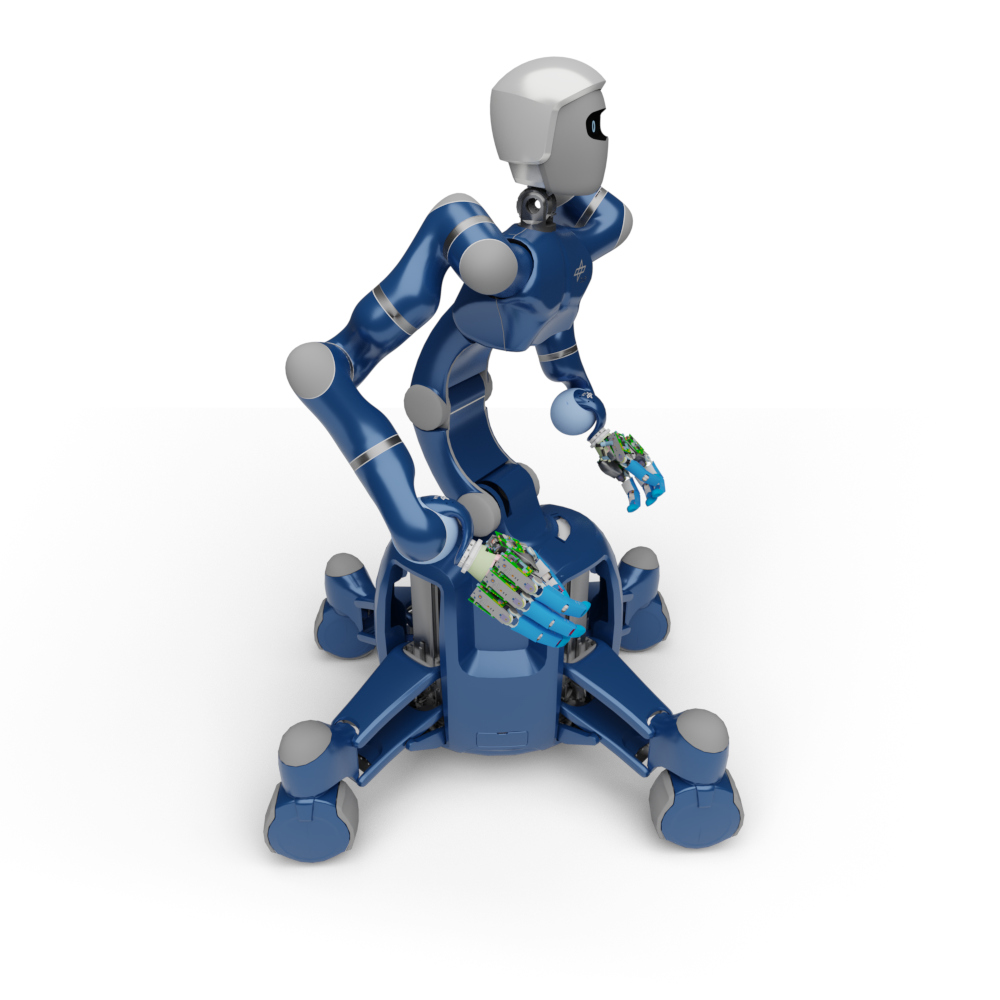}};
\set{\babyImageSize}{0.65*\imageSize};
\node [inner sep=0pt, anchor=south] at (\xJustinPos - \diffBetweenJustins *0.935, \yJustinLower + \imageSize *0.1) {\includegraphics[width=\babyImageSize cm]{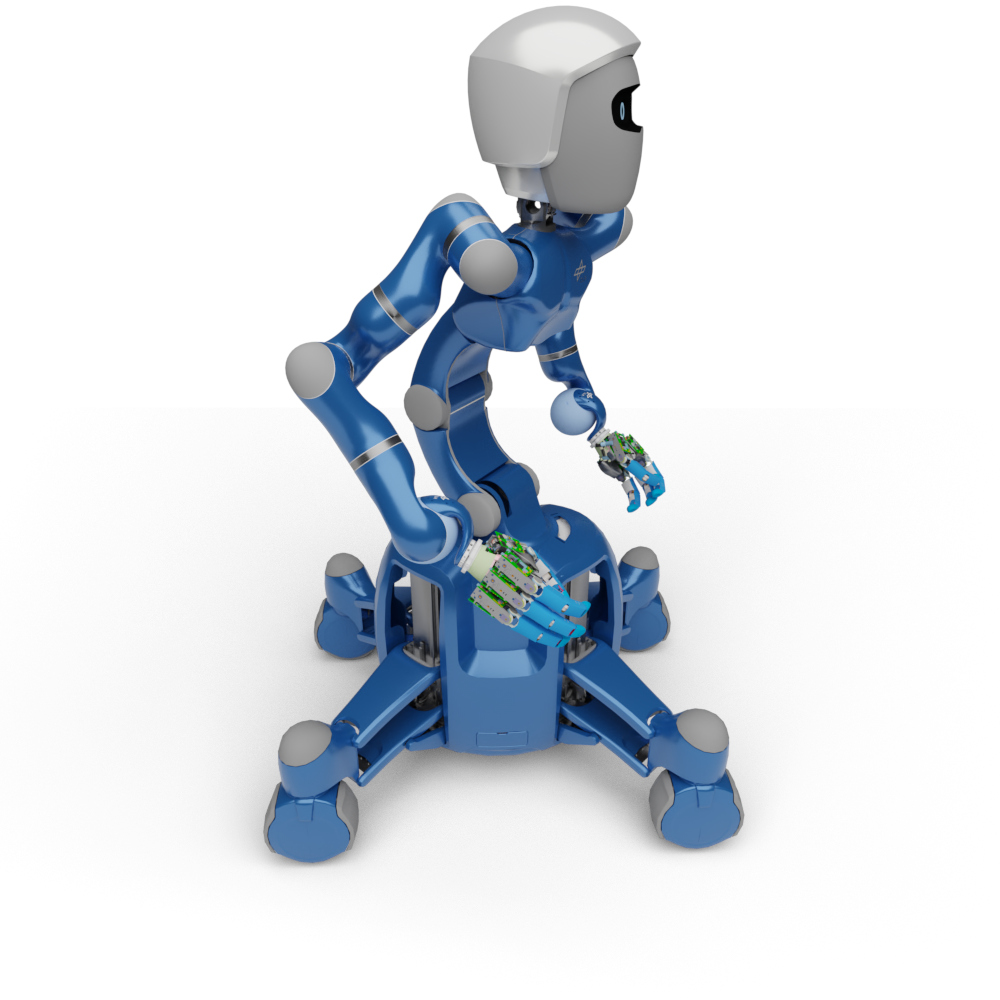}};
\node [inner sep=0pt,  anchor=south] at (\xJustinPos + \diffBetweenJustins, \yJustinLower) {\includegraphics[width=\imageSize cm]{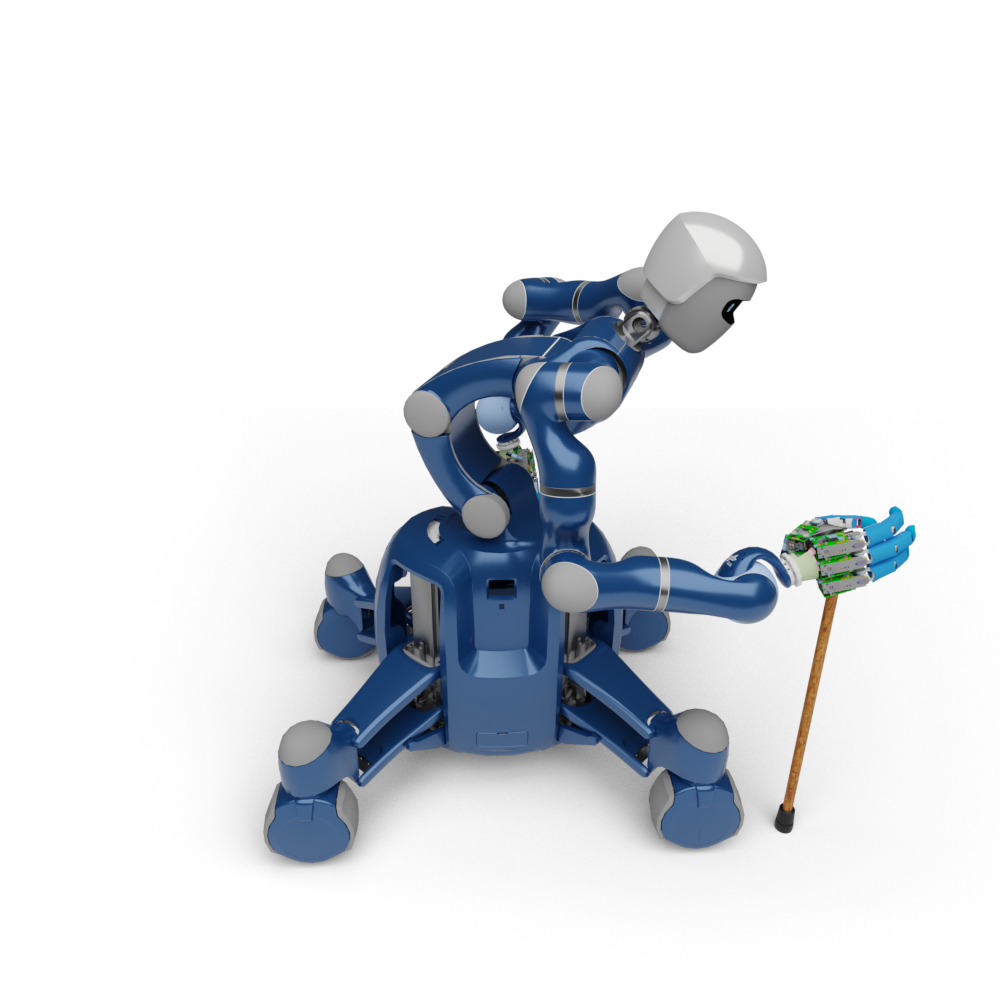}};
\draw [->] (\xJustinPos - \diffBetweenJustins*0.6, \yJustinLower + \imageSize * 0.5) -> (\xJustinPos - \diffBetweenJustins*0.6 + 0.2, \yJustinLower + \imageSize * 0.5);
\draw [->] (\xJustinPos + \diffBetweenJustins*0.45, \yJustinLower + \imageSize * 0.5) -> (\xJustinPos + \diffBetweenJustins*0.45 + 0.2, \yJustinLower + \imageSize * 0.5);
\node [inner sep=0pt, anchor=south, align=center] at (\xstart + \xSeqFirstPos, \yJustinLower-0.05) {\scriptsize Lifelong learning};

\end{tikzpicture}
}
\caption{Continual learning without using previously seen examples. In each sequence, the model has to learn new categories while also ensuring not to forget previous ones.}
\label{fig:introPicture}
\end{figure}

%% file: chapters/relatedWork.tex
Continual learning can be divided mainly into architectural, regularization, rehearsal, and a combination of those different strategies.

Progressive Neural Networks (PNN), by Rusu \etal \cite{Rusu2016}, and Copy Weight with Reinit (CWR), by Lomonaco \etal \cite{Lomonaco2017}, solve the problem of catastrophic forgetting by an \textbf{architectural strategy}. 
Both works let their network dynamically grow to accommodate new categories.
However, in PNN, new sub-networks are repeatedly connected to the old frozen sub-networks, ensuring the usage of prior knowledge, while in CWR, new weights are trained for new sequences, which are then frozen after training.
In contrast to that, in our work, all weights are trained at the same time, enabling the network to learn those differences and prevent it from forgetting.

In literature, learning without forgetting (LwF) \cite{Li2017} is often used for comparison \cite{Maltoni2019, Rebuffi2017, Zenke2017, LopezPaz2017, Stojanov2019}.
They propose a \textbf{regularization strategy} to stabilize the model accuracy on old tasks using knowledge distillation, proposed by Hinton \etal \cite{Hinton2015}.
Here, the logits of the previous and current sequences are encouraged to be similar when applied to data from the new sequence. 
Further, they propose a form of knowledge distillation, where they record a set of label probabilities for each training image on the previous network weights.
In LWF, they compensate for the distribution shift in the predicted probability distribution of different sequences by first freezing the old neurons and only training the new neurons in a warm-up step. 
Whereas in our work, we record the logits directly instead of modeling them through a probability distribution. 
Further, we propose to use a regression loss to compensate for distribution shifts, as it removes them completely.

Another regularization strategy is Elastic Weight Consolidation (EWC) \cite{Kirkpatrick2017}, proposed by Kirkpatrick \etal, which tries to protect important weights from changing by using a Fisher weight importance matrix.
Another one is Synaptic Intelligence (SI) \cite{Zenke2017}, by Zenke \etal, where they propose calculating the weight importance on the fly, using Stochastic Gradient Descent.
Both approaches are rooted in neural science, as they argue that a biological synapse accumulates task-relevant information over time and stores new memories without forgetting old ones.
In this work, we change all weights on purpose to give the network the ability to learn all kinds of separations, and we tackle the forgetting by enforcing that the network's outputs don't change.

The most similar approach to our work is AR1 by Maltoni and Lomonaco \cite{Maltoni2019}.
It is a \textbf{combination} of an architectural strategy (CWR+) and a regularization strategy (SI).
AR1 uses importance weighting and therefore needs the entire backbone during training.
They also zero-init their last layer and propose to use mean-shifting to improve their continual learning performance. 
In \OURAPPROACH, on the other hand, only the backbone is frozen and used to generate the features beforehand, thus decreasing training time. 
Furthermore, each sequence gets its own head to improve the overall capacity of the network and its accuracy instead of just one single layer in AR1.
Lastly, AR1 uses a classification loss instead of a regression loss. 

All previously mentioned approaches are rehearsal-free and therefore comparable to ours, as no examples of previous tasks are available here. 
Another way of continual learning is given by \textbf{rehearsal strategies}, where previous training examples or representations of them are stored and used in later training steps \cite{Belouadah2019, Ayub2020, Ayub2021}.
Two relevant methods are Incremental Classifier and Representation Learning (\iCaRL), by Rebuffi \etal \cite{Rebuffi2017}, and Gradient Episodic Memory (GEM) from Lopez-Patz and Ranzato \cite{LopezPaz2017}.
In \iCaRL, they propose a class-incremental algorithm to learn new classes over time. 
It relies on a regularization strategy in a nearest-mean-of-exemplar classification and a rehearsal strategy by saving feature representations in each sequence.
Like LWF, \iCaRL uses a combination of knowledge distillation and classification loss to train the network.
GEM uses the same strategies, but while \iCaRL is designed to fill the whole memory after every batch, GEM uses a fixed amount of memory for each batch, a so-called episodic memory. Instead of keeping the predictions of past sequences invariant by using distillation, GEM uses the losses as an inequality constraint.
This avoids their increase but allows their decrease in order to make the positive backward transfer possible.  
Like Rebuffi \etal \cite{Rebuffi2017}, our approach focuses on keeping the predictions of past sequences invariant and does not consider the possibility of a positive backward transfer proposed in GEM. 
Another approach is Persistent Anytime Learning of Objects from Unseen Classes (PAL) by Denninger \etal \cite{Denninger2018}.
Similar to us, they also use CNNs for feature extraction, but instead of fully connected layers, they use a random forest classifier, which also relies on saving previously seen data samples.
In their online learning scenario, they found that the removal of trees leads to catastrophic forgetting and that the performance of learning new categories decreases over time.

These approaches reduce forgetting by saving samples from previous sequences. 
However, our overarching goal in this work is to get closer to a system that is able to learn continually without depending on storing previously seen examples.
A lifelong learning strategy should not contain the requirement to store everything it has ever seen.

%% file: chapters/approach.tex
This chapter introduces \OURAPPROACH, our algorithm for rehearsal-free object classification on 2D images for continual learning. 
We selected ResNet50 as a feature extractor and backbone, which is pre-trained on Imagenet \cite{He2016, Deng2009}.

\subsection{Architectural Strategies}\label{architecturalStrategies}

Due to the continual procedure, where new categories are shown in sequences, we propose to let the network grow in each sequence. This is depicted in \cref{fig:introPicture} and \cref{fig:reconstructionLabel}.
Here, the first sequence is learned as in a classical CNN. 
For each following sequence, we propose to add a new head.
Each newly added head has two fully-connected layers, called $\layerName{\sequence}{0}$ and $\layerName{\sequence}{1}$ with the sequence number $\sequence$.
Whereas each $\layerName{\sequence}{1}$ has the same amount of outputs as categories $\classes_{\sequence}$ are in sequence $\sequence$.
During the training of this new sequence, all weights of all added heads are trained, and only the backbone is kept frozen.
The input to each head is the same $2048$ feature vector.
By doing this, the network can better address the feature separation in specific sequences. 
This also avoids that the training of the categories of a new sequence changes any of the weights of prior sequences heads.

\subsection{Regularization Strategies}\label{regularizationStrategies}

The next step is to tackle catastrophic forgetting, which describes the decreasing performance on the categories of prior sequences when new sequences are trained. 
Our goal is now to prevent the logits of the categories of previous sequences from changing.

\subsubsection{Recall label}\label{reconstructionLabel}

\input{tikz_figures/reconstructionLabel}

In the first sequence, image-label pairs $(\images, \olabels)$ are used for training, where the label $\olabel$ represents the one-hot encoded category of an object.
In the following sequences, however, these one-hot encoded categories might lead to catastrophic forgetting, as the ground truth labels for categories, which are no longer present, are zero. 
To reach this zero value during training, the categories logit values of the network have to continually move into the negative, as they can only reach zero in the softmax if $x \to -\infty$ in $e^{x}$.
This then changes the previously learned categories by forcing them to predict zeros regardless of the input.
The logit value here is the input to the final softmax.
In order to prevent this forgetting, we propose another label for already learned categories.
As a label represents a value, which the network's output should target, we propose to use the original logit output on the training image.
We call this value \reconstructionLabelWord $\reconstructionlabel$. 
This is similar to the approach from Li \etal \cite{Li2017}, but they save the final softmax output and not the logit itself.
We argue that the softmax output already has too much influence from the other classes making the decoupling harder.

But, the labels $\olabel_{\prevSequence}$ from any previous sequence $\prevSequence$ are no longer present in the next sequence $\sequence$. 
Thus, we only have the image-label pairs $(\images_\sequence, \olabels_\sequence)$.
Therefore, we propose replacing all zeros in the current $\olabel_{\sequence}$, which stand for the previous categories' labels with our \reconstructionLabelWord $\reconstructionlabel_\sequence$.
This \reconstructionLabelWord $\reconstructionlabel_\sequence$ is then compared to the output of $\layerName{0}{1}$ at the current sequence $\sequence$.
The training procedure for this is shown in \cref{algo:TrainingProcedure}, where the calculation of the recall label happens in line three. 
This procedure is also depicted in \cref{fig:reconstructionLabel}.

Given the model $\network$ from our method, sequences $\sequences \subseteq \mathbb{N}_0$ and categories $\classes \subseteq \mathbb{N}_0$, where each sequence $\sequence \in \sequences$ has its own categories $\classes_{\sequence} \subseteq \classes$, where $\classes_{s_i} \cap \classes_{s_j} = \emptyset$ and $s_i \neq s_j$ for $s_i,s_j \in \sequences$ and its network weights $\weights_{\sequence}$, we do:

\begin{itemize}
	\item In the first step, the model $\networkWithoutX$ with its weights $\weights_{0}$ is trained using the image-class pairs $(\image_0, \olabel_0)$ of sequence $\sequence = 0$ (see \cref{algo:TrainingProcedure}, line six).
	\item For each new sequence $\sequence \in \sequences$, the \reconstructionLabelWordPlural $\reconstructionlabel_{\sequence}$ are calculated for each image of the current sequence $\sequence$, using the weights $\weights_{\prevSequence}$ of the previous sequence (see \cref{algo:TrainingProcedure}, line three).
	\item We create new network weights $\weights_{\sequence}$ by adding a new head $\networkWithoutX_{\prevSequence}$ ($\layerName{1}{0}$, $\layerName{1}{1}$) according to the number of categories in the current sequence $|\classes_\sequence|$.
	\item Now, the model $\networkWithoutX_{\sequence}$ can be trained. 
\end{itemize}

\begin{algorithm}[h]
	\begin{algorithmic}[1]
		\small
		\Procedure{train\_network}{Network $\networkWithoutX$, Sequence $\sequence$, Categories $\classes$, Images $\image_{\sequence}$, Labels $\olabel_{\sequence}$}
		
		\If{$\sequence$ $\neq$ $0$}
		\State {$\reconstructionlabel_{\sequence} = \left[ \networkWithoutX.\text{predictWithoutSoftmax}(X = \trainingsexample) \text{ for } i \text{ in } x \right]$}
		\State $\olabel_{\sequence} = \text{concatenate}\left(\reconstructionlabel_{\sequence},\olabel_{\sequence}\right)$
		\EndIf
		\State {$\textsc{expand\_network}(\networkWithoutX,	\vert(\classes_{\sequence})\vert)$}		
		\State {$\networkWithoutX.\text{train}(X=\image_\sequence, Y=\olabel_{\sequence})$}
		\State \Return {$\network$}
		\EndProcedure
	\end{algorithmic}
	\caption[Network training procedure]{Network training procedure}
	\label{algo:TrainingProcedure}
\end{algorithm}

\subsubsection{Loss function}\label{lossFunction}

In order to enable the network to learn how to reconstruct the previous network outputs and simultaneously learn new categories, we have to adapt the loss function.
So, we define a new loss, as we are now using \reconstructionLabelWordPlural $\reconstructionlabel_{\sequence}$ for the categories of previous sequences and one-hot encoded classification labels for the current categories.
Our loss function consists out of three parts:
\paragraph{1. Loss on the previous categories $\bm{\lossonoldweights}$}
The aim is to force the logit output space of the categories of previous sequences $\classknown$ to only slightly adapt to the categories of the current sequence $\classes_{\sequence}$, but to mainly stay the same, as those previous categories are not represented in the training anymore.
	\begin{equation}\label{eq_squaredDifferenceRecon}
	\lossonoldweights\trainingsexampleAndSequence = \frac{1}{|\classknown|} \sum_{\class\, \in \, \classknown} \left(\networkoutput\forTrainingExampleAndClass - \reconstructionlabel\forTrainingExampleAndClass\right)^{2}
	\end{equation}
	\Cref{eq_squaredDifferenceRecon} shows the loss on previous categories $\lossonoldweights\trainingsexampleAndSequence$, where a regression loss is used instead of a classification loss.
	Namely, an L2-norm is calculated between the \reconstructionLabelWord $\reconstructionlabel$ and the network's logit output $\networkoutput$ for a given training example $\trainingsexample$ and a given category $\class$. This is done for all $\class$ in $\classknown$: 	
\begin{equation}\label{eq_squaredDifferenceReconcomb}
	\classknown = \begin{cases}
	\emptyset & \text{if }\sequence = 0\\
	\bigcup\limits_{\sequence'}^{[0, \dots ,\prevSequence]} \classes_{\sequence'} & \text{else}
	\end{cases}
\end{equation}
	With this loss function, previously shown categories are now prevented from being forgotten. 
	Nevertheless, it is still necessary to ensure that the network learns new categories, which is tackled next.
	
\paragraph{2. Loss on the current categories $\bm{\lossonnewweights}$:} 
	In order to learn the categories of the current sequence, a cross-entropy loss with softmax is used on the newly generated heads, which is possible as classification labels are used in a one-hot encoding style for the current sequence. The loss on new categories for one training example in the current sequence $\lossonnewweights$, can be described as:  
	
	\begin{equation}\label{eq_lossonnewweights}
	\lossonnewweights\trainingsexampleAndSequence = \frac{1}{|\classes_\sequence|} \sum_{\class \, \in \, \classes_\sequence} \binaryIndicator\forTrainingExampleAndClass \log(\predictionPropability\forTrainingExampleAndClass)
	\end{equation}
	With this loss $\lossonnewweights$, our network is able to learn new categories, where $\predictionPropability\! \left[\class\right]$ is the softmax output for a class $\class$.
	To balance those two-loss parts, a third loss is defined next.  
	
\paragraph{3. Loss over all categories $\bm{\lossonallweights}$:} 
	As the goal is to have a complete classifier for all categories, we now have to find a solution for the decoupling of the previous categories from the new ones.
	Here, also the cross entropy loss with softmax as in \cref{eq_lossonnewweights} is used, but this time on all logit outputs (all previous and new categories) $\classes'_{\sequence} = \bigcup\nolimits_{\sequence'}^{[0,\dots,\sequence]} \classes_{\sequence'}$, where $\sequence$ is the current sequence:
	\begin{equation}\label{eq_lossonallweights}
	\lossonallweights\trainingsexampleAndSequence = \frac{1}{|\classes'_{\sequence}|} \sum_{\class  \, \in \, \classes'_{\sequence}} \binaryIndicator\forTrainingExampleAndClass \log(\predictionPropability\forTrainingExampleAndClass)
	\end{equation}
	Important to note here is that this loss part again uses classification labels and a classification loss.
	Therefore, this loss function tries to lower the logit response of previous categories, as they are represented with a zero value in the labels, which might lead to forgetting. 
	The recall loss $\lossonoldweights$ prevents this.
Combining the three formula parts, our loss function on all training examples of the current sequence $\sequence$ is defined in \cref{eq_loss}.
The loss in the first sequence only uses \cref{eq_lossonnewweights} of the loss function as there are no previous categories given.

\begin{equation}\label{eq_loss}
\loss_{\sequence} =
\begin{cases}
\lossonnewweights\trainingsexampleAndSequence & \text{if } \sequence = 0 \\
 \lossonoldweights\trainingsexampleAndSequence + \lossonnewweights\trainingsexampleAndSequence + \lossonallweights\trainingsexampleAndSequence & \text{else}
\end{cases}	 
\end{equation}

\subsubsection{Discrepancy in the output distribution}\label{discrepancy}

Continually training the network sequence after sequence causes the logit outputs $\networkoutput$ to rise steadily, which leads to forgetting. 
See the blue line in \cref{fig:variancePerSequenceAndMethod}.
This is caused as the previous categories try to keep their logit values through the $\lossonoldweights$, which means that to learn new categories, the logit values of those categories have to be higher than before. 
As the softmax uses all categories, this then increases the variance of each sequence.
Caused through this unequal comparison inside of the softmax, the older sequences categories will be increasingly forgotten over time.
We solve this problem of distribution discrepancy within already learned categories by dividing the difference between the output and the \reconstructionLabelWord by the variance per category.

\begin{equation}\label{eq_varianzPerClass}
\varianz[\class] = \mathbb{E} [(\networkoutput[\class] - \mean[\class])^2], \forall \class \in \classknown
\end{equation}
In \cref{eq_varianzPerClass}, the calculation of the variance per category $\varianz[\class]$ is shown, where $\mean[\class]$ is the mean over the logit outputs $\networkoutput$ during one sequence $\sequence$ over all the image-label pairs $(\images_\sequence, \olabels_\sequence)$ for one class $\class$.
So, the loss on the previous categories from \cref{eq_squaredDifferenceRecon} is changed, resulting in a modified loss $\lossonoldweightsvariance$, shown in \cref{eq_varianz}.
Inspired by a Mahalanobis distance, this equation shows the network output difference $\networkoutput\forTrainingExampleAndClass$ and the recall label $\reconstructionlabel\forTrainingExampleAndClass$ of each neuron being divided by the variance of the respective category $\varianz[\class]$.
This reduces the discrepancy in the output of the neurons.

\begin{equation}\label{eq_varianz}
\lossonoldweightsvariance\trainingsexampleAndSequence = \frac{1}{|\classknown|} \sum_{\class\, \in \, \classknown} \left(\frac{\networkoutput\forTrainingExampleAndClass - \reconstructionlabel\forTrainingExampleAndClass}{\varianz[\class]}\right)^{2}
\end{equation}
The red line in \cref{fig:variancePerSequenceAndMethod} shows that this technique smooths the variance.

\input{tikz_figures/variance.tex}

\subsubsection{Full regression loss}\label{reconstructionLoss}

The division by the variance, however, does not solve the problem entirely as it only tries to patch up the problems caused by the softmax and cross-entropy.

As this is the root cause, we propose to replace the softmax and cross-entropy with a regression loss and clamp the output of the last layer to the range of zero to one, which is shown in \cref{eq_clampedOutput}.

\begin{equation}\label{eq_clampedOutput}
\networkoutputClipped\forTrainingExampleAndClass = \text{max}(0, \text{min}(\networkoutput\forTrainingExampleAndClass, 1))
\end{equation}

After that, we need to adapt $\lossonnewweights$ and $\lossonallweights$ by replacing the cross entropy with an L2 loss, see \cref{eq_regression_loss} and \cref{eq_regression_loss_2}.

\begin{align}\label{eq_regression_loss}
	\reconstructionLossNewWeights\trainingsexampleAndSequence &= \tfrac{1}{|\classes_\sequence|} \sum_{\class \, \in \, \classes_\sequence}\parenth{\networkoutputClipped\forTrainingExampleAndClass - \binaryIndicator\forTrainingExampleAndClass}^2 \\ \label{eq_regression_loss_2}
	\reconstructionLossAllWeights\trainingsexampleAndSequence &= \tfrac{1}{|\classes'_{\sequence}|} \sum_{\class  \, \in \, \classes'_{\sequence}} \parenth{\networkoutputClipped\forTrainingExampleAndClass - \binaryIndicator\forTrainingExampleAndClass}^2
\end{align}

Resulting in $\reconstructionLossNewWeights$ and $\reconstructionLossAllWeights$, which are now used in \cref{eq_loss}.
The black line in \cref{fig:variancePerSequenceAndMethod} depicts that our proposed \reconstructionLossName smooths the output distribution best.
Nevertheless, we lose the advantage that the sum of all values is one.
Finally, we also try using the variance here to shift the focus to categories, which have a high range of values in their corresponding logit output.

%% file: tikz_figures/reconstructionLabel.tex
\begin{figure}
\centering
\resizebox{\columnwidth}{!}{
	\begin{tikzpicture}
	
	\set{\xStart}{0};
	\set{\yStart}{0};
	\set{\imageSize}{2.25};
	\set{\spacing}{0.7};
	\set{\filterWidth}{0.7};
	\set{\filterHeight}{3};
	\set{\arrowSpacing}{0.1};
	\set{\filterAmount}{2};
	\set{\classAmount}{5};
	\set{\secondClassAmount}{3};
	\set{\extractionWidth}{1.8};
	\set{\boxHeight}{4.5};
	\set{\boxWidth}{4}
	\set{\boxSpacing}{1.2};
	
	\drawPlane{\xStart}{-(\yStart+\imageSize)*0.5}{0}{\imageSize}{\imageSize}{generalStyle, fill=white, draw=black, thick};
	
	\node[draw, shape=circle, anchor=base] at (\xStart, \yStart+\boxHeight/2) {1.};
	\node[text ragged, align=center, anchor=north west] at (\xStart+0.35, \yStart+\boxHeight/2+0.4) {Prediction of the \reconstructionLabelWordPlural $\reconstructionlabel_\sequence$ before training a new sequence.};
	
	\node [inner sep=0pt] at (\xStart + \imageSize * 0.5, \yStart) {\includegraphics[width=\imageSize cm]{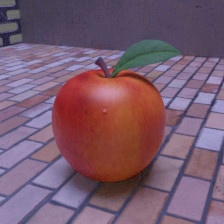}};
	\node [text ragged, align=center, anchor=north, yshift=-9pt] at (\xStart + \imageSize * 0.5, \yStart - \imageSize * 0.5) {Training images of\\ current sequence $\sequence$\\ (224$\times$224$\times$3)};

	\set{\xExtractionStart}{\xStart + \imageSize + \spacing * 0.5};
	
	\drawIndividualFrustum{\xExtractionStart}{-(\yStart+\imageSize)*0.5}{0}{\imageSize}{\filterHeight*0.3}{\extractionWidth}{generalStyle, rounded corners=0.3mm, fill=hellblue};
	
	\node[text ragged, align=center] at (\xExtractionStart+\extractionWidth*0.5,0) {frozen \\ ResNet50};
	\node[text ragged, align=center, anchor=north, yshift=-3pt]  at (\xExtractionStart+\extractionWidth*0.5, \yStart - \extractionWidth) {Output \\ (2048)};
	
	\set{\filterStart}{\xExtractionStart + \extractionWidth};
	
	\set{\lowerYPos}{-(\yStart+\filterHeight)*0.5};
	\newcommand{\nrNeurons}{{2048,\classAmount}};
	\drawPlane{\filterStart+\spacing/2}{\lowerYPos - \boxSpacing}{0}{\boxWidth}{\boxHeight}{generalStyle, fill=white, draw=black, rounded corners=1mm};
	\foreach \layerNr in {1, ..., \filterAmount}{
		\set{\xPos}{\layerNr * \spacing + (\layerNr - 1) * \filterWidth + \filterStart};
		\set{\lastXPos}{(\layerNr - 1) * \spacing + (\layerNr - 2) * \filterWidth +\filterStart};
		
		\def\fillColor{lightgray};
		\ifthenelse{\layerNr = 2}{\def\fillColor{baseClasses}}{};
		\drawPlane{\xPos}{\lowerYPos}{0}{\filterWidth}{\filterHeight}{generalStyle, draw=black, rounded corners=1mm, fill=\fillColor};
		\pgfmathparse{\nrNeurons[\layerNr - 1]}
		\set{\nrCurrentNeurons}{\pgfmathresult};
		\set{\layerNrMinusOne}{int(\layerNr-1)};
		\node [text ragged, align=center, anchor=north,yshift=-3pt] at (\xPos + \filterWidth*0.5, \lowerYPos) {\layerNameBroken{0}{\layerNrMinusOne}{\nrCurrentNeurons}{}};
		
		\draw [->, thick] (\lastXPos + \filterWidth + \arrowSpacing, \yStart) -- (\xPos - \arrowSpacing, \yStart);
	
	}
	
	\set{\classAmountMinusOne}{\classAmount - 1};
	\foreach \layerNr in {1, ..., \classAmountMinusOne}{
		\set{\xPos}{\filterAmount * \spacing + (\filterAmount  - 1) * \filterWidth + \filterStart};
		\set{\xPosEnd}{\xPos + \filterWidth};
		\set{\yPos}{\lowerYPos + (\filterHeight / \classAmount) * \layerNr};	
		
		\draw (\xPos, \yPos) -- (\xPosEnd, \yPos);
		
	}
	
	\node[text ragged, align=center, anchor=north,yshift=-3pt] at (\filterStart+\spacing/2+\boxWidth/2, \lowerYPos - \boxSpacing) {Weights of the \\ previous network $\networkWithoutX_{\prevSequence}$};
	
	\set{\xPos}{\filterAmount * \spacing +\arrowSpacing + (\filterAmount  - 1) * \filterWidth + \filterStart + \filterWidth};
	\draw [->, thick] (\xPos, \yStart) -- (\xPos + \spacing - \arrowSpacing, \yStart);
	\node [text ragged, align=center, anchor=west] at (\xPos + \spacing - \arrowSpacing, \yStart) (networkOutput) {$\reconstructionlabel_{\sequence}$};
	
	\set{\sumClasses}{int(\classAmount+\secondClassAmount)}
	\set{\filterAmount}{1};
	
	\set{\yStartprev}{\yStart}
	\set{\yStart}{\yStartprev - \boxHeight - \boxSpacing*1.7};
	
	\drawPlane{\xStart}{\yStart-\imageSize*0.5}{0}{\imageSize}{\imageSize}{generalStyle, fill=white, draw=black, thick};
	
	\node[draw, shape=circle, anchor=base] at (\xStart, \yStart+\boxHeight/2) {2.};
	\node[text ragged, align=center, anchor=north west] at (\xStart+0.35, \yStart+\boxHeight/2+0.4) {Training of the current sequence after expanding the network.};
	
	\node [inner sep=0pt] at (\xStart + \imageSize * 0.5, \yStart) {\includegraphics[width=\imageSize cm]{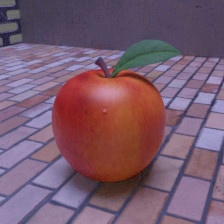}};
	\node [text ragged, align=center, anchor=north, yshift=-9pt] at (\xStart + \imageSize * 0.5, \yStart - \imageSize * 0.5) {Training images of\\ current sequence $\sequence$\\ (224$\times$224$\times$3)};

	\set{\xExtractionStart}{\xStart + \imageSize + \spacing * 0.5};
	
	\drawIndividualFrustum{\xExtractionStart}{\yStart- \imageSize*0.5}{0}{\imageSize}{\filterHeight*0.3}{\extractionWidth}{generalStyle, rounded corners=0.3mm, fill=hellblue};
	\set{\yFrustumMiddle}{\yStart};
	
	\set{\xBackBoneEnd}{\xExtractionStart+\extractionWidth};
	
	\node[text ragged, align=center] at (\xExtractionStart+\extractionWidth*0.5,\yStart) {frozen \\ ResNet50};
	\node[text ragged, align=center, anchor=north, yshift=-3pt]  at (\xExtractionStart+\extractionWidth*0.5, \yStart - \extractionWidth) {Output \\ (2048)};
	
	\set{\filterStart}{\xExtractionStart + \extractionWidth};
	
	\set{\lowerYPos}{\yStart -\filterHeight*0.5};
	\drawPlane{\filterStart+\spacing/2}{\lowerYPos - \boxSpacing}{0}{\boxWidth}{\boxHeight}{generalStyle, fill=white, draw=black, rounded corners=1mm};
	\set{\yBoxMiddle}{\lowerYPos - \boxSpacing +\boxHeight*0.5};
	
	\set{\distBetween}{0.64};
	\set{\filterBoxFactor}{0.825};
	\set{\spacingResNet}{0.2};
	\set{\filterHeight}{\filterBoxFactor * \boxHeight};
	\set{\filterWidth}{\filterWidth * 0.8};
	\set{\spacing}{0.5 / 0.8};
	\set{\spacingHalf}{\spacing*0.5};
	\set{\filterStart}{\filterStart+\spacing/2+\arrowSpacing};
	\set{\boxWidth}{\spacing*4.2+\filterAmount*\filterWidth+\extractionWidth+\imageSize+\spacingResNet};
	\set{\lowerYPos}{\yBoxMiddle - \boxHeight / 2 + (\boxHeight - \filterHeight) / 2 + 0.25};
	
	\newcommand{\nrCategories}{{\secondClassAmount, \classAmount}};
	\set{\sumClasses}{\classAmount + \secondClassAmount};
	\set{\filterAmount}{2};
	\foreach \layerNr in {1, ..., \filterAmount}{
		\set{\xPos}{\layerNr * \spacing + (\layerNr - 1) * \filterWidth + \filterStart};
			
		\ifthenelse{\layerNr = 2}{\def\fillColor{baseClasses}}{};
		\ifthenelse{\layerNr = 2}{	
			\foreach \partNr in {0, ..., 1}{
				\pgfmathparse{\nrCategories[\partNr]}
				\set{\nrCurrentNeurons}{\pgfmathresult};			
				\ifthenelse{\partNr = 0}{
					\def\fillColor{sequenceClasses}
					\set{\ySize}{\nrCurrentNeurons / \sumClasses * (\filterHeight - \distBetween)};
					\set{\yOffset}{((\filterHeight *0.5 - \distBetween * 0.5) + \ySize) / 2 - \ySize};
					\set{\yTextOffset}{0}
					\set{\currentClass}{\secondClassAmount};
				}{
					\def\fillColor{baseClasses};
					\set{\yOffset}{0.5 * \filterHeight + \distBetween * 0.5};
					\set{\ySize}{0.5 * (\filterHeight - \distBetween)};
					\set{\yTextOffset}{\filterHeight*0.5+\distBetween*0.5}
					\set{\currentClass}{\classAmount};
				};
				\drawPlane{\xPos}{\lowerYPos+ \yOffset}{0}{\filterWidth}{\ySize}{imgStyle, draw=black, rounded corners=0.75mm, fill=\fillColor};		
				\set{\partNrCorrected}{int(2 + \partNr*-1 - 1)};
				
				\set{\layerNrMinusOne}{int(\layerNr-1)};
				\node [text ragged, align=center, anchor=north,yshift=-(\partNr - 1)* 0.3pt + 1.5pt, font=\scriptsize\linespread{0.8}\selectfont] at (\xPos+\filterWidth/2, \lowerYPos+\yTextOffset) {\scriptsize  \layerNameBroken{\partNrCorrected}{\layerNrMinusOne}{\currentClass}{\scriptsize}};
			}
		}{
			\drawPlane{\xPos}{\lowerYPos+\filterHeight*0.5+\distBetween*0.5}{0}{\filterWidth}{\filterHeight*0.5-\distBetween*0.5}{imgStyle, draw=black, rounded corners=0.75mm, fill=lightgray};
			\drawPlane{\xPos}{\lowerYPos}{0}{\filterWidth}{\filterHeight*0.5-\distBetween*0.5}{imgStyle, draw=black, rounded corners=0.75mm, fill=lightgray};
			
			\set{\layerNrMinusOne}{int(\layerNr-1)};
			\node [text ragged, align=center, anchor=north,yshift=1.5pt, font=\scriptsize\linespread{0.8}\selectfont] at (\xPos+\filterWidth/2, \lowerYPos) {\scriptsize \layerNameBroken{1}{\layerNrMinusOne}{2048}{\scriptsize}};
			\node [text ragged, align=center, anchor=north,yshift=1.5pt, font=\scriptsize\linespread{0.8}\selectfont] at (\xPos+\filterWidth/2, \lowerYPos+\filterHeight*0.5+\distBetween*0.5) {\scriptsize \layerNameBroken{0}{\layerNrMinusOne}{2048}{\scriptsize}};
			
		}	
		\set{\yLowerMiddle}{\lowerYPos+(\filterHeight*0.5-\distBetween*0.5)*0.5};
		\set{\yUpperMiddle}{\lowerYPos+\filterHeight-(\filterHeight*0.5-\distBetween*0.5)*0.5};
	
		\ifthenelse{\layerNr = 1}{
			\set{\lastXPos}{\xBackBoneEnd};		
			\set{\yUpperStart}{\yFrustumMiddle};	
			\set{\yLowerStart}{\yFrustumMiddle};
		}{
			\set{\lastXPos}{\xPos - \spacing};	
			\set{\yUpperStart}{\yUpperMiddle};	
			\set{\yLowerStart}{\yLowerMiddle};
		}
	
		\draw [thick, ->] (\lastXPos + \arrowSpacing, \yLowerStart)  to [out=0,in=180] (\xPos - \arrowSpacing, \yLowerMiddle);
		\draw [thick, ->] (\lastXPos + \arrowSpacing, \yUpperStart)  to [out=0,in=180] (\xPos - \arrowSpacing, \yUpperMiddle);
		
	}
	\set{\xPos}{\filterStart+\filterWidth+\spacing*2};
	
	\foreach \partNr in {0, ..., 1}{
		\pgfmathparse{\nrCategories[\partNr]}
		\set{\nrCurrentNeurons}{\pgfmathresult};		
		\set{\classAmountMinusOne}{\nrCurrentNeurons - 1}
		\set{\currentFilterHeight}{\nrCurrentNeurons / \sumClasses * \filterHeight};
		\ifthenelse{\partNr = 0}{
			\def\fillColor{sequenceClasses}
			\set{\ySize}{\nrCurrentNeurons / \sumClasses * (\filterHeight - \distBetween)};
			\set{\yOffset}{((\filterHeight *0.5 - \distBetween * 0.5) + \ySize) / 2 - \ySize};
			\draw [decorate, decoration={brace, mirror, amplitude=10pt},xshift=1pt] (\xPos + \filterWidth, \lowerYPos+\yOffset) -- (\xPos + \filterWidth, \lowerYPos+ \ySize+\yOffset) node [text ragged, align=center, anchor=west, midway, xshift=8pt] (lossOnNewWeights) {$\lossonnewweights_{\sequence}$};
			
			\node [right= of lossOnNewWeights, anchor=west] (yLabel) {$\olabel_\sequence$};
			\draw [->, thick] (yLabel) -> (lossOnNewWeights);
		}{
			\def\fillColor{baseClasses};
			\set{\yOffset}{0.5 * \filterHeight + \distBetween * 0.5};
			\set{\ySize}{0.5 * (\filterHeight - \distBetween)};
			\draw [decorate, decoration={brace, mirror, amplitude=10pt},xshift=1pt] (\xPos + \filterWidth, \lowerYPos+\yOffset) -- (\xPos + \filterWidth, \lowerYPos+ \ySize+\yOffset) node [text ragged, align=center, anchor=west, midway, xshift=8pt] (lossOnOldWeights) {$\lossonoldweights_{\sequence}$};
	
		};
		\foreach \currentlayerNr in {1, ..., \classAmountMinusOne}{
			\set{\xPosEnd}{\xPos + \filterWidth};
			
			\set{\yPos}{\lowerYPos + \yOffset + (\ySize / \nrCurrentNeurons) * \currentlayerNr};	
			\draw (\xPos, \yPos) -- (\xPosEnd, \yPos);
		}
	}
	
	\draw [->, thick] (networkOutput.east) to [out=305,in=55] node[line width=0.4pt, xshift=3pt, yshift=-5pt, left, midway, anchor=west] {Use $\reconstructionlabel_{\sequence}$} (lossOnOldWeights.east);

	\end{tikzpicture}
}
\caption[Calculation of the recall label]{Calculation of the \reconstructionLabelWord $\reconstructionlabel$: First, we calculate the logits for each training example of the current sequence $\sequence$.
After that, a new head is added ($\layerName{1}{0}$, $\layerName{1}{1}$) and the \reconstructionLabelWordPlural $\reconstructionlabel_{\sequence}$, are used for training in sequence $\sequence$.
}
\label{fig:reconstructionLabel}
\end{figure}

%% file: tikz_figures/variance.tex
\begin{figure}
	\resizebox{0.95\columnwidth}{!}{
	\pgfplotstableread[col sep = comma,]{Plots/variance_new.csv}\datatable
	\begin{tikzpicture}
	\begin{axis}[
	log ticks with fixed point, ytick={12, 25, 60},
	ymode=log, xticklabels={}, name=first plot, height=4.5cm, width=\columnwidth,
	legend style={nodes={scale=0.7, transform shape},column sep=5pt, at={(0.02, 0.98)}, anchor=north west, legend columns=1,fill=none},
	]
	\addplot+ [mark options={fill=sequenceClasses}, sequenceClasses] table[x expr=\coordindex, y={Standard}]{\datatable};
	\addlegendentry{RECALL}
	\addplot+ [mark options={fill=baseClasses, scale=1.3}, baseClasses, mark=diamond*] table[x expr=\coordindex, y={Dividing by variance}]{\datatable};
	\addlegendentry{RECALL var.}
	\end{axis}
	\begin{axis}[
	at=(first plot.south), anchor=north,
	log ticks with fixed point, ytick={0.3, 1, 3},
	ymode=log, , xlabel={Sequences}, height=4.5cm, width=\columnwidth,
	legend style={nodes={scale=0.7, transform shape},column sep=5pt, at={(0.98, 0.5)}, anchor=east, legend columns=1,},
	]
	\addplot [mark options={fill=black}, draw=black, mark=square*] table[x expr=\coordindex, y={Reconstruction}]{\datatable};
	\addlegendentry{RECALL reg.}
		\addplot [mark options={fill=sicolor}, draw=sicolor, mark=square*] table[x expr=\coordindex, y={ReconstructionDivVar}]{\datatable};
	\addlegendentry{RECALL var. reg.}

	\end{axis}
	\end{tikzpicture}
	}
	\caption{Logarithmic variance over all logits per sequence and method. In blue, the output variance for the standard \OURAPPROACH with catastrophic forgetting is shown. In red \OURAPPROACH using dividing by variance, in black \OURAPPROACH using \reconstructionLossName and in green \OURAPPROACH using \reconstructionLossName in combination with dividing by variance is depicted. \coreFifty \cite{Lomonaco2017} is used here.}
	\label{fig:variancePerSequenceAndMethod}
\end{figure}

%% file: chapters/howsDataset.tex
In order to better show the strength of our approach, we present a novel synthetic dataset for object classification for continual learning, created with BlenderProc, called \textbf{\DLRDataset} (\textbf{H}ousehold \textbf{O}bjects \textbf{W}ithin \textbf{S}imulation dataset for \textbf{C}ontinual \textbf{L}earning).
BlenderProc, by Denninger \etal \cite{Denninger2020}, is a procedural pipeline to generate images for deep learning.  
Our dataset contains 150,795 unique synthetic images using 25 different household categories with 925 3D models in total, see \cref{fig:myDataset}.
We achieved that by creating a room with randomly textured floors, walls, and a light source with randomly chosen light intensity and color.
After that, a 3D model is placed in the resulting room. 
This object gets customized by randomly assigning materials including different textures in order to achieve a diverse dataset.
Moreover, each object might be deformed with a random displacement texture.
For each RGB-D image, we also provide the corresponding segmentation map and normal image.

The images are organized in five sequences, containing five categories each.
We also provide a long-version with twelve sequences containing three categories in the first and two in the following sequences.
At the end, ten percent of the images are used for validation, whereas an object instance can either be in the training or in the validation set. 
This avoids that the network learns to recognize instances of certain categories.
We created this dataset by taking 774 3D models from the ShapeNet dataset \cite{Chang2015} and the others models from various sites.
The dataset and the code are available online: \url{https://github.com/DLR-RM/RECALL}.

\input{tikz_figures/MyDataset}

\subsection{Dataset comparison}

\input{tables/datasets}

In our opinion, one of the most relevant applications for online learning is robotics, where a dataset consisting of household objects is more relevant than out of cars, planes, boats and dogs.
This is the most important feature setting \DLRDataset apart from \iCIFAR and ImageNet1K, making it similar to \coreFifty but with a pronounced focus on category than on instance learning.
Compared with \coreFifty, our dataset contains two and a half times more categories and over 18.5 times more instances, as well as a wider variety of backgrounds, lighting conditions, and camera positions.
Furthermore, our dataset is non-handheld, as a CNN could learn the category of the object based on the grasp, which is not ideal if, after training, the approach is used in a general setting \cite{Wang2017}.
In \cref{tab_datasets}, we show datasets that other continual learning papers have used. 
This table shows that we provide the widest variety of objects and sessions from all datasets, and we are the only ones providing a segmentation map and normal image to each RGB-D image.
Here, a session is a specific environment.

Rebuffi \etal \cite{Rebuffi2017} introduce another dataset that is often used in continual learning, called \iCIFAR, which is an incremental version of the CIFAR-100 dataset proposed by Krizhevsky \etal \cite{Krizhevsky2009} with 100 categories.
This dataset splits the categories equally into different task sets (sequences).
The advantage of this dataset is that it contains more categories than \coreFifty or our dataset, but the problem is that it is easier for a network, which is pre-trained on ImageNet \cite{Deng2009} as those contain similar or even the same categories.
On top of that, it is not a household dataset.

OpenLORIS is a robotic vision dataset proposed by She \etal \cite{She2020}, which contains household objects recorded by a robot. 
The difference in our dataset is that OpenLORIS focus on instance learning, as each sequence contains the same 69 objects but in different conditions (occlusion, view change, and more), and additionally, this dataset uses the same instances and sessions for training and testing.  

One of the best advantages of a synthetic dataset is that it is only a question of computation time to create an even more extensive dataset, whereas expanding the number of categories or instances in \coreFifty is much more time-intensive, as these images are recorded manually. 
One might be concerned that using synthetic data will not generalize to real-world images.
But, as shown from Hoda{\v{n}} \etal \cite{Hodan2019} and Denninger \etal \cite{Denninger2020}, a generalization from synthetic to real-world images is possible.

%% file: tikz_figures/MyDataset.tex
\begin{figure}
\centering
\resizebox{0.9625\columnwidth}{!}{
	\begin{tikzpicture}
	
	\set{\xStart}{0};
	\set{\yStart}{0};
	\set{\imageAmount}{24};
	\set{\imageSize}{3};
	\set{\spacing}{0.1};
	\set{\arrowSpacing}{0.1};
	\set{\classAmount}{10};
	\set{\extractionWidth}{1.8};

	\foreach \image in {0, ..., \imageAmount}{
		\pgfmathparse{\image - int(\image / 5) * 5} 
		\set{\imageMod}{\pgfmathresult};
		\set{\yPos}{\xStart - (\spacing+\imageSize) * int(\image / 5)};
		\set{\xPos}{\yStart + (\spacing+\imageSize) * \imageMod};
		\node [inner sep=0pt] at (\xPos, \yPos) {\includegraphics[width=\imageSize cm]{pictures/new_classes/class_\image.jpg} };
	}
	\end{tikzpicture}
}
\caption{All 25 categories used in the \DLRDataset dataset are suited for mobile robotics as shown here. For training and testing, different instances of these categories are used.}
\label{fig:myDataset}
\end{figure}

%% file: tables/datasets.tex
\begin{table*}
	\centering 
	\caption{Compared to other datasets used for continual learning, \DLRDataset has the most objects, sessions and sequences.}
	\label{tab_datasets}
	{
	\resizebox{1\textwidth}{!}{		
	\begin{tabular}{|l|*{8}{c|}}\hline 
		\textbf{Dataset} & \textbf{Type} & \textbf{Imgs.} & \textbf{Cat.} & \textbf{Obj.} & \textbf{Sess.} & \textbf{Seq.} & \textbf{Format} & \textbf{Setting}\\ \hline 
		permuted MNIST \cite{LeCun1998, Kirkpatrick2017} & Category  & $70000$ & $10$ & - & $1$ & $10$ & grayscale & hand written\\ 
		OpenLORIS \cite{She2020} & Instance & $\bm{1106424}$ & $19$ & $69$ & $7$ & $9$ & RGB-D & robot camera \\ \hline
		\coreFifty \cite{Lomonaco2017} & Instance & $164866$ & $10$ & $50$ & $11$ & $9$ & RGB-D & hand hold\\ 	
		\iCIFAR \cite{Rebuffi2017} & Category & $60000$ & $\bm{100}$ & $600$ & $600$ & $10$ & RGB & mixed\\ \hline \hline
		\DLRDataset & Category  & $150795$ & $25$ & $\bm{925}$ & $\bm{50265}$ & $5$ or $\bm{12}$ & \textbf{RGB-D, segmap, normal} & synthetic \\ \hline
	\end{tabular}
	}
	}
\end{table*}

%% file: chapters/evaluation.tex
\subsection{Experimental setup}

We evaluate our approach on \iCIFAR created by Rebuffi \etal \cite{Rebuffi2017}, on \coreFifty, proposed by Lomonaco \etal \cite{Lomonaco2017}, and finally on both versions of \DLRDataset, our dataset created with BlenderProc. 
In addition to the rehearsal-free methods (AR1, LwF, EWC, SI), we also compare our approach with \iCaRL and A-GEM (see \cref{relatedWork}) except for the \coreFifty dataset, which is not supported out of the box for those approaches.
All evaluated approaches use the same features generated by a frozen ResNet50 backbone on RGB input data.
This ResNet50 is pre-trained on ImageNet.
The only exception is AR1 as this approach actively changes the backbone while training, so using the features is impossible.
For AR1, we use the code provided by the authors, which uses a MobileNet as backbone  \cite{Howard2017}.
To improve comparability we also tested our approach with the same MobileNet backbone, shown in \cref{tab_ablation_studies}.
For generating the results on LwF, EWC, SI, iCaRL and A-GEM we use code provided by Van de Ven \etal \cite{VandeVen2019} and adapt it so that all approaches us the same features.

\input{tables/resultComparison}

\input{tikz_figures/allResults}

\subsection{Discussion}

\input{tables/ablation_studies}
The results of our experiments are depicted in \cref{tab_result_comparison} and \cref{fig:results}.
Our approach reaches the best accuracy of all rehearsal-free methods with $71.45\%$ on \coreFifty,  as shown in the top plot of \cref{fig:results}. \OURAPPROACH even outperforms AR1, which is to the best of our knowledge state-of-the-art for rehearsal-free continual learning on this dataset.

Based on the results, it is quite clear that LwF suffers from catastrophic forgetting.
The distribution shift in LwF still seems to be a problem, as the network forgets quite fast even though they use the final output of the softmax as target values.
\OURAPPROACH solves this by using \reconstructionLabelWordPlural $\reconstructionlabel_{\sequence}$ and a full regression loss.
The bad performance of SI and EWC shows that only using an importance matrix seems insufficient. 
AR1, which combines SI with an architectural strategy, improves the performance by a decent margin.

On \iCIFAR, \OURAPPROACH reaches $61.15\%$ in contrast to $42.39\%$ with AR1, even though the implementation provided by the authors of AR1 reaches a better performance on \iCIFAR than the approx. $31\%$ reported in their paper \cite{Maltoni2019}.
In comparison to the other rehearsal-free learning approaches, we are able to learn new categories without strong forgetting.

On \DLRDataset, our approach is also the best performing rehearsal-free method with $57.83\%$ by a big margin, see third plot of \cref{fig:results}.
For the long version of \DLRDataset, the best performing \OURAPPROACH version is the one with \reconstructionLossName and dividing by variance with $40.65\%$ compared to the standard one with $36.85\%$.
This shows that replacing the probability distribution in the \reconstructionLossName mode works better on more complex tasks.
However, in such a scenario with more sequences and complex categories, the limit of our approach is shown, as forgetting can't be prevented entirely, see the last plot in \cref{fig:results}.

The experiments on the \DLRDataset dataset show the advantage of rehearsal strategies, where training data is kept for later sequences.
As \DLRDataset contains more objects than \coreFifty or \iCIFAR, which seems to be harder for all rehearsal-free methods.
This observation is even more vital for the challenging long version of \DLRDataset, see the bottom plot of \cref{fig:results}.
Here, it is shown that each rehearsal-free method forgets almost everything in the last sequence.
On less challenging datasets like \iCIFAR, \OURAPPROACH even performs better than the rehearsal strategies, see the second plot of \cref{fig:results}. 
It is interesting how \iCaRL performs on this dataset, as it first suffers from forgetting but is able to relearn most of the categories in the last sequence. 
This is only possible because it saves previously seen training examples. 
Remarkably, \OURAPPROACH performs almost as well as the rehearsal strategies on both versions of \DLRDataset, which shows that using the \reconstructionLabelWordPlural $\reconstructionlabel_{\sequence}$ helps the network to recall previous categories without using any memory.

The contrast in performance between \OURAPPROACH and the other methods gets even bigger if one focuses solely on categorical continual learning tasks. 
Here, the method has to understand the semantic meaning of an object rather than recognizing one specific instance.
This is visible for AR1, which is not able to learn the categories of \DLRDataset, as the category instances used for testing are different from the ones used for training.
By testing \OURAPPROACH with MobileNet features we find that this is not caused by the used backbone of AR1, see \cref{tab_ablation_studies}.
For \coreFifty, where the instances stay the same, the performance difference of AR1 is smaller.

In addition, \OURAPPROACH is faster than AR1 as we can directly work on the backbone features. These are saved in a TFRecord file, so one complete run (training and validation) takes roughly three minutes\footnotemark[2].
The one-time conversion of all training images of \coreFifty to ResNet50 features takes roughly 6 minutes\footnotemark[2]. 
In comparison, AR1 adapts the backbone and takes roughly 55 min.\ for each run of \coreFifty. 
For detailed sequence analysis, we refer to the appendix.

\footnotetext[1]{This value is from \cite{Maltoni2019}, in our experiments we reach $57.67\%$.}
\footnotetext[2]{Using a NVIDIA GeForce RTX 2080 Ti}

\subsection{Ablation}\label{sec_ablation}

As \OURAPPROACH is fast to train, it was possible to do several ablation studies.
In the following, we present hyperparameters, which have a strong influence on the results.

\paragraph{Feature extraction network}

We evaluated the influence of the backbone on the performance, see \cref{tab_ablation_studies}.
We use ResNet50 as our default backbone in order to achieve better comparability over all datasets, even though InceptionResNetV2, proposed by Szegedy \etal \cite{Szegedy2017}, performs better on the \DLRDataset dataset.
As \DLRDataset is a category classification dataset, we assume that more advanced CNNs have an improved accuracy in contrast to \coreFifty, which is an instance-classification dataset.
It can also be seen that ResNet50 proposed by He \etal \cite{He2016} works better than the second version ResNet50V2 also proposed by He \etal \cite{He2016a}.
On the \coreFifty dataset, ResNet50 outperforms every other tested backbone.
Other papers like Maltoni \etal \cite{Maltoni2019} have partly confirmed these results, where they also found that ResNet50 outperformed GoogLeNet, proposed by Szegedy \etal \cite{Szegedy2015} on the \coreFifty dataset.

\paragraph{Using different heads per sequence}

In \cref{tab_ablation_studies} it can be seen that using a different head per sequence instead of only changing the last layer improves the results.
Especially for the \coreFifty dataset, this can be explained by the fact that separating different sequences reduces the possible influence of new output values on the first layer during training.

\paragraph{Activation function}

In our approach, different activation functions are tested, see \cref{tab_ablation_studies}.
SIREN proposed by Sitzmann \etal \cite{Sitzmann2020} performs best for both datasets.
As SIREN has been successfully used in reconstruction tasks, it is interesting to see that it also works best in a classification scenario \cite{Nair2010}. 
But, as highlighted in their paper, it highly depends on its hyperparameter $\omega_0$.
In the case of the standard version of \DLRDataset, ReLU \cite{Nair2010} performs better.

%% file: tables/resultComparison.tex
\begin{table*}
	\centering 
	\caption{\OURAPPROACH performs best on various datasets, compared to other rehearsal-free approaches,  measured by the accuracy over all categories after the last sequence. This average accuracy and std. deviation is calculated over 40 runs.}
	\label{tab_result_comparison}
	{
		\small
	\begin{tabular}{|l|c|c|c|c|c|c|c|}
		\hline
		Approach/Dataset & 	\coreFifty (instance) & \iCIFAR (category) & \DLRDataset (category) & \DLRDataset long (category) \\ \hline
		LwF \cite{Li2017} & $34.14$ & $27.93$ & $25.13$ & $10.29$ \\
		EWC \cite{Kirkpatrick2017} & $43.29$ & $9.67$ & $17.99$ & $7.01$ \\
		SI \cite{Zenke2017} & $26.77$ & $9.62$ & $16.25$ & $6.96$ \\
		AR1 \cite{Maltoni2019} & $69.48$\footnotemark & $42.39$ & $8.59$ & $8.41$ \\ \hline
		\OURAPPROACH & $64.57 (\pm 0.79)$ & $\bm{61.15 (\pm 0.29)}$ & $\bm{57.83 (\pm 0.18)}$ & $36.85 (\pm 0.23)$ \\
		\OURAPPROACH var. & $50.36 (\pm 1.06)$ & $61.09 (\pm 0.31)$ & $55.68 (\pm 1.27)$ & $36.48 (\pm 0.22)$ \\
		\OURAPPROACH reg. & $63.13 (\pm 2.46)$ & $56.52(\pm 0.23)$ & $57.05(\pm 0.53)$ & $39.19(\pm 1.67)$ \\
		\makecell{\OURAPPROACH var. reg.} & $\bm{71.45 (\pm 0.43)}$ & $56.46 (\pm 0.24)$ & $56.82 (\pm 0.95)$ & $\bm{40.65 (\pm 1.37)}$ \\

		\hline 
	\end{tabular}
	}
\end{table*}

%% file: tikz_figures/allResults.tex
\set{\plotSize}{0.84}
\begin{figure}
	\centering
	\begin{subfigure}[b]{\columnwidth}
		\centering
		\resizebox{\plotSize\columnwidth}{!}{	
			\begin{tikzpicture}
			\begin{axis}[
			title={Results on \coreFifty},
			title style={at={(-0.3,0.5)},anchor=north, rotate=90},
			legend style={column sep=5pt, at={(0.98, 0.98)}, anchor=north east, nodes={scale=0.8, transform shape}},
			ymin=0,
			ymax=100,
			ylabel={Accuracy in \%}]
			\addplot+ [aronestyle] table[x=Sequence, y=AR1-Ours, col sep=comma] {Plots/core50.csv};
			\addlegendentry{AR1 results (from \cite{Maltoni2019})}
			\plotGraph{Plots/core50.csv}{0}{0}{1}
			\end{axis}
			\end{tikzpicture}
		}
	\end{subfigure}
	\begin{subfigure}[b]{\columnwidth}
		\centering
		\moveUpInImage
		\resizebox{\plotSize\columnwidth}{!}{	
			\begin{tikzpicture}
			\begin{axis}[
			title={Results on \iCIFAR},
			title style={at={(-0.3,0.5)},anchor=north, rotate=90},
			legend style={legend pos=outer north east},
			ymin=0,
			ymax=100,
			ylabel={Accuracy in \%}]
			\plotGraph{Plots/cifar100.csv}{0}{1}{1}
			\end{axis}
			\end{tikzpicture}
		}
	\end{subfigure}
	\begin{subfigure}[b]{\columnwidth}
	\centering
	\moveUpInImage
	\resizebox{\plotSize\columnwidth}{!}{	
		\begin{tikzpicture}
		\begin{axis}[
		title={Results on \DLRDataset},
		title style={at={(-0.3,0.5)},anchor=north, rotate=90},
		legend style={legend pos=outer north east},
		ymin=0,
		ymax=100,
		ylabel={Accuracy in \%}]
		\plotGraph{Plots/hows.csv}{0}{1}{1}
		\end{axis}
		\end{tikzpicture}
	}
	\end{subfigure}
	\begin{subfigure}[b]{\columnwidth}
		\centering
		\moveUpInImage
		\resizebox{\plotSize\columnwidth}{!}{	
			\begin{tikzpicture}
			\begin{axis}[
			title={Results on \DLRDataset long version},
			title style={at={(-0.3,0.5)},anchor=north, rotate=90},
			legend style={legend pos=outer north east},
			xlabel={Sequence},
			ymin=0,
			ymax=100,
			ylabel={Accuracy in \%}]
			\plotGraph{Plots/hows-long.csv}{0}{1}{1}
			\end{axis}
			\end{tikzpicture}
		}
	\end{subfigure} 
	\begin{subfigure}[t]{0.5\textwidth}
		\centering
		\begin{tikzpicture}
			\begin{axis}[	
				legend columns=4,
				legend style={column sep=5pt, at={(0.98, 0.98)}, anchor=north east, nodes={scale=0.8, transform shape}},
				hide axis,
				xmin=0,
				xmax=12,
				ymin=0,
				ymax=100]
				\plotGraph{Plots/core50.csv}{1}{1}{0}
			\end{axis}
		\end{tikzpicture}
	\end{subfigure}
	\caption{Results on four datasets. \iCaRL and A-GEM are not rehearsal-free and are here to show how strong our approach is. \OURAPPROACH is SOTA for rehearsal-free learning in all plots. We always use the best run here.For the exact values, see the appendix.}
	\label{fig:results}
\end{figure}

%% file: tables/ablation_studies.tex
\begin{table*}
	\centering
	\caption{The accuracies over all categories after the last sequence are depicted for various feature-extractors, strategies, and activation functions. The used hyperparameters in \OURAPPROACH are marked with a star.}
	\label{tab_ablation_studies}
	{
	\resizebox{1\textwidth}{!}{	
	\begin{tabular}{| l | c | c | c | c | c || c | c || c | c |} \hline
		\multirow{3}{*}{Dataset} & \multicolumn{5}{c||}{Feature-Extractor} & \multicolumn{2}{c||}{Strategy} & \multicolumn{2}{c|}{Activation function} \\ \cline{2-10}
		& Mobile- & ResNet50* & ResNet50V2 & Inceptionv3 & Inception- & Exp. last &  Adding & ReLU & SIREN* \\ 
		& Net\cite{Howard2017} & \cite{He2016} & \cite{He2016a} & \cite{Szegedy2016} & ResNetV2 \cite{Szegedy2017} & layer &  head* & \cite{Nair2010} & \cite{Sitzmann2020} \\ \hline
		\coreFifty & $71.85$ & $\bm{72.00}$ & $64.04$ & $61.87$ & $51.55$ & $64.44$ & $\bm{72.00}$ & $59.35$ & $\bm{72.00}$\\ \hline
		\DLRDataset & $44.47$ & $59.17$ & $59.51$ & $67.66$ & $\bm{69.19}$ & $55.15$ & $\bm{58.31}$ & $\bm{59.17}$ & $58.31$\\
		\hline
	\end{tabular}
	}
	}
\end{table*}

%% file: chapters/conclusion.tex
This paper shows that the challenging task of online learning can be efficiently solved by combining several architectural and regularization techniques.
We demonstrate SOTA performance on the two datasets \coreFifty and \iCIFAR, in a rehearsal-free setting.
Our approach \OURAPPROACH is able to adapt to new categories by adding a new head per sequence.
In order to prevent forgetting, we introduce \reconstructionLabelWordPlural. 
However, the usage of those might lead to a discrepancy in the output distribution, for which we propose two different solutions.
First, we propose to normalize the logits outputs by dividing with the variance per category and show top performance on \iCIFAR. 
In the second solution, we replace the classification with a regression and outperform AR1 on \coreFifty.
Further, we present a novel dataset for continual learning, especially suited for object recognition in a mobile robot environment, called \DLRDataset.
On this, we show a strong improvement in comparison to all other rehearsal-free learning methods.

%% file: tikz_figures/MyDatasetApple.tex
\begin{figure}
\centering
\resizebox{0.9\columnwidth}{!}{
	\begin{tikzpicture}
	
	\set{\xStart}{0};
	\set{\yStart}{0};
	\set{\imageAmount}{3};
	\set{\imageSize}{3.75};
	\set{\spacing}{0.1};
	\set{\arrowSpacing}{0.1};
	\set{\extractionWidth}{1.8};

	\foreach \image in {0, ..., \imageAmount}{
		
		\pgfmathparse{\image - int(\image / 5) * 5} 
		\set{\imageMod}{\pgfmathresult};
		\set{\yPos}{\xStart - (\spacing+\imageSize) * int(\image / 5)};
		\set{\xPos}{\yStart + (\spacing+\imageSize) * \imageMod};
		\node [inner sep=0pt] at (\xPos, \yPos) {\includegraphics[width=\imageSize cm]{pictures/apple/apple_\image.jpg} };
	}
	\end{tikzpicture}
}
\caption{Each RGB image of an object in the \DLRDataset dataset (left) has a corresponding segmentation map (second from left), a normal image (third from left), and a depth image (right).}
\label{fig:myDatasetApple}
\end{figure}

%% file: tikz_figures/hows_standard.tex
\begin{figure*}
\centering
\resizebox{0.998\textwidth}{!}{
	\begin{tikzpicture}
	
	\set{\xstart}{0}
	\set{\ystart}{0}
	\set{\lineLength}{16}
	\set{\imageSize}{1}
	\set{\arrowSpacing}{0.02};
	\set{\classAmount}{5};
	\set{\nrSequences}{5}
	\set{\bufferleft}{1}
	\set{\labelSpace}{(\lineLength-\bufferleft-\bufferleft)/(\nrSequences-1)} 
	
	\node[inner sep=0pt] (lineStart) at (\xstart, \ystart) {};
	\node[inner sep=0pt] (lineEnd) at (\xstart + \lineLength, \ystart) {};
	
	\draw[->, thick] (lineStart) -> (lineEnd);
	\node [right= 0.05cm of lineEnd] {time};
	
	\set{\xstartfirstSeq}{\xstart + \bufferleft}
	\node [circle, draw, inner sep=1.5pt, fill] at (\xstartfirstSeq, \ystart) (firstCircle) {};
	\node [anchor=north, below= 0.1cm of firstCircle] (seqzerotext) {Seq $0$};
	
	\set{\xstartsecSeq}{\xstartfirstSeq + \labelSpace}
	\node [circle, draw, inner sep=1.5pt, fill] at (\xstartsecSeq, \ystart) (secondCircle) {};
	\node [anchor=north, below= 0.1cm of secondCircle] (seqonetext) {Seq $1$};
	
	\set{\xstartthirdSeq}{\xstartsecSeq + \labelSpace}
	\node [circle, draw, inner sep=1.5pt, fill] at (\xstartthirdSeq, \ystart) (thirdCircle) {};
	\node [anchor=north, below= 0.1cm of thirdCircle] (seqtwotext) {Seq $2$};
	
	\set{\xstartfourthSeq}{\xstartthirdSeq + \labelSpace}
	\node [circle, draw, inner sep=1.5pt, fill] at (\xstartfourthSeq, \ystart) (fourthCircle) {};
	\node [anchor=north, below= 0.1cm of fourthCircle] (seqtreetext) {Seq $3$};
	
	\set{\xstartfifthSeq}{\xstartfourthSeq + \labelSpace}
	\node [circle, draw, inner sep=1.5pt, fill] at (\xstartfifthSeq, \ystart) (fifthCircle) {};
	\node [anchor=north, below= 0.1cm of fifthCircle] (seqfourtext) {Seq $4$};

	\node [anchor=north, right= 0.1cm of seqfourtext] {$\hdots$};

	\node [inner sep=0pt, above= 0.15cm of firstCircle, anchor=south] (firstImgFirst)  {\includegraphics[width=\imageSize cm]{pictures/new_classes/class_3.jpg}};
	\node [imgStyle, inner sep=0pt, above right= 0.1cm and 0.05cm of firstImgFirst, anchor=south] (firstImgSecond) {\includegraphics[width=\imageSize cm]{pictures/new_classes/class_1.jpg}};
	\node [imgStyle, inner sep=0pt, above left= 0.1cm and 0.05cm of firstImgFirst, anchor=south] (firstImgThird) {\includegraphics[width=\imageSize cm]{pictures/new_classes/class_2.jpg}};
	\node [imgStyle, inner sep=0pt, left= 0.1cm of firstImgFirst, anchor=east] (firstImgFourth) {\includegraphics[width=\imageSize cm]{pictures/new_classes/class_0.jpg}};
	\node [imgStyle, inner sep=0pt, right= 0.1cm of firstImgFirst, anchor=west] (firstImgFivth) {\includegraphics[width=\imageSize cm]{pictures/new_classes/class_4.jpg}};
	
	\node [inner sep=0pt, above= 0.15cm of secondCircle, anchor=south] (secondImgFirst) 
	{\includegraphics[width=\imageSize cm]{pictures/new_classes/class_5.jpg}};
	\node [imgStyle, inner sep=0pt, above right= 0.1cm and 0.05cm of secondImgFirst, anchor=south] (firstImgSecond) {\includegraphics[width=\imageSize cm]{pictures/new_classes/class_6.jpg}};
	\node [imgStyle, inner sep=0pt, above left= 0.1cm and 0.05cm of secondImgFirst, anchor=south] (firstImgThird) {\includegraphics[width=\imageSize cm]{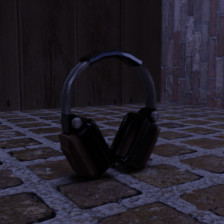}};
	\node [imgStyle, inner sep=0pt, left= 0.1cm of secondImgFirst, anchor=east] (firstImgFourth) {\includegraphics[width=\imageSize cm]{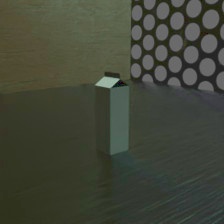}};
	\node [imgStyle, inner sep=0pt, right= 0.1cm of secondImgFirst, anchor=west] (firstImgFivth) {\includegraphics[width=\imageSize cm]{pictures/new_classes/class_9.jpg}};
	
	\node [inner sep=0pt, above= 0.15cm of thirdCircle, anchor=south] (thirdImgFirst) 
	{\includegraphics[width=\imageSize cm]{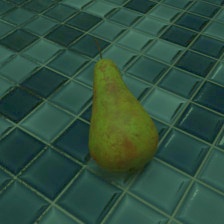}};
	\node [imgStyle, inner sep=0pt, above right= 0.1cm and 0.05cm of thirdImgFirst, anchor=south] (firstImgSecond) {\includegraphics[width=\imageSize cm]{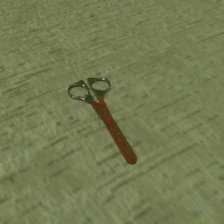}};
	\node [imgStyle, inner sep=0pt, above left= 0.1cm and 0.05cm of thirdImgFirst, anchor=south] (firstImgThird) {\includegraphics[width=\imageSize cm]{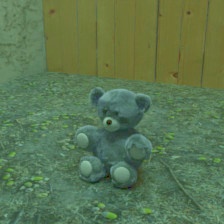}};
	\node [imgStyle, inner sep=0pt, left= 0.1cm of thirdImgFirst, anchor=east] (firstImgFourth) {\includegraphics[width=\imageSize cm]{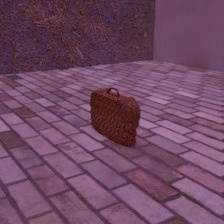}};
	\node [imgStyle, inner sep=0pt, right= 0.1cm of thirdImgFirst, anchor=west] (firstImgFivth) {\includegraphics[width=\imageSize cm]{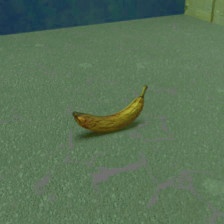}};
	
	\node [inner sep=0pt, above= 0.15cm of fourthCircle, anchor=south] (fourthImgFirst) 
	{\includegraphics[width=\imageSize cm]{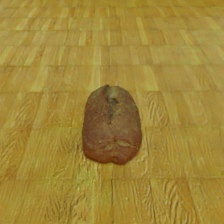}};
	\node [imgStyle, inner sep=0pt, above right= 0.1cm and 0.05cm of fourthImgFirst, anchor=south] (firstImgSecond) {\includegraphics[width=\imageSize cm]{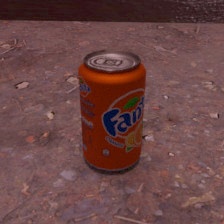}};
	\node [imgStyle, inner sep=0pt, above left= 0.1cm and 0.05cm of fourthImgFirst, anchor=south] (firstImgThird) {\includegraphics[width=\imageSize cm]{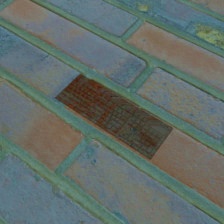}};
	\node [imgStyle, inner sep=0pt, left= 0.1cm of fourthImgFirst, anchor=east] (firstImgFourth) {\includegraphics[width=\imageSize cm]{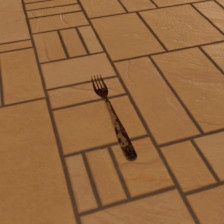}};
	\node [imgStyle, inner sep=0pt, right= 0.1cm of fourthImgFirst, anchor=west] (firstImgFivth) {\includegraphics[width=\imageSize cm]{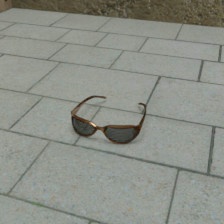}};
	
	\node [inner sep=0pt, above= 0.15cm of fifthCircle, anchor=south] (fifthImgFirst) 
	{\includegraphics[width=\imageSize cm]{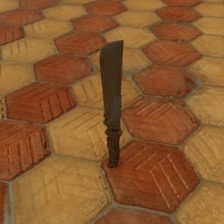}};
	\node [imgStyle, inner sep=0pt, above right= 0.1cm and 0.05cm of fifthImgFirst, anchor=south] (firstImgSecond) {\includegraphics[width=\imageSize cm]{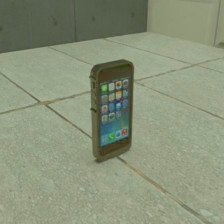}};
	\node [imgStyle, inner sep=0pt, above left= 0.1cm and 0.05cm of fifthImgFirst, anchor=south] (firstImgThird) {\includegraphics[width=\imageSize cm]{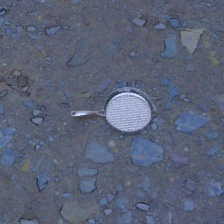}};
	\node [imgStyle, inner sep=0pt, left= 0.1cm of fifthImgFirst, anchor=east] (firstImgFourth) {\includegraphics[width=\imageSize cm]{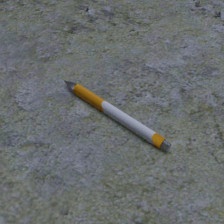}};
	\node [imgStyle, inner sep=0pt, right= 0.1cm of fifthImgFirst, anchor=west] (firstImgFivth) {\includegraphics[width=\imageSize cm]{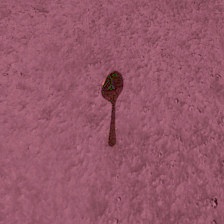}};

	\end{tikzpicture}
}
\caption{Standard version of \DLRDataset dataset with five sequences.}
\label{fig:howsStandard}
\end{figure*}

%% file: tikz_figures/hows_long.tex
\begin{figure*}
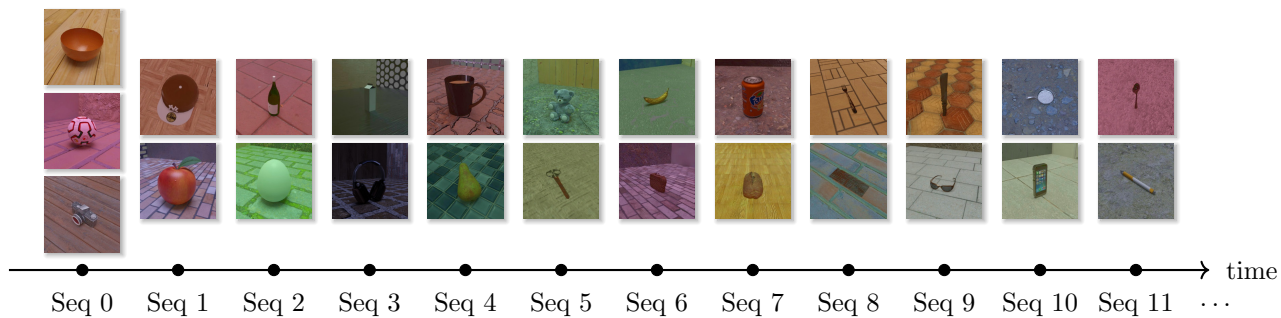

\centering
	\begin{tikzpicture}
	
	\set{\xstart}{0}
	\set{\ystart}{0}
	\set{\lineLength}{16}
	\set{\imageSize}{1}
	\set{\arrowSpacing}{0.02};
	\set{\classAmount}{5};
	\set{\nrSequences}{12}
	\set{\bufferleft}{1}
	\set{\labelSpace}{(\lineLength-\bufferleft-\bufferleft)/(\nrSequences-1)} 
	
	\node[inner sep=0pt] (lineStart) at (\xstart, \ystart) {};
	\node[inner sep=0pt] (lineEnd) at (\xstart + \lineLength, \ystart) {};
	
	\draw[->, thick] (lineStart) -> (lineEnd);
	\node [right= 0.05cm of lineEnd] {time};
	
	\set{\xstartfirstSeq}{\xstart + \bufferleft}
	\node [circle, draw, inner sep=1.5pt, fill] at (\xstartfirstSeq, \ystart) (firstCircle) {};
	\node [anchor=north, below= 0.1cm of firstCircle] (seqzerotext) {Seq $0$};
	
	\set{\xstartsecSeq}{\xstartfirstSeq + \labelSpace}
	\node [circle, draw, inner sep=1.5pt, fill] at (\xstartsecSeq, \ystart) (secondCircle) {};
	\node [anchor=north, below= 0.1cm of secondCircle] (seqonetext) {Seq $1$};
	
	\set{\xstartthirdSeq}{\xstartsecSeq + \labelSpace}
	\node [circle, draw, inner sep=1.5pt, fill] at (\xstartthirdSeq, \ystart) (thirdCircle) {};
	\node [anchor=north, below= 0.1cm of thirdCircle] (seqtwotext) {Seq $2$};
	
	\set{\xstartfourthSeq}{\xstartthirdSeq + \labelSpace}
	\node [circle, draw, inner sep=1.5pt, fill] at (\xstartfourthSeq, \ystart) (fourthCircle) {};
	\node [anchor=north, below= 0.1cm of fourthCircle] (seqtreetext) {Seq $3$};
	
	\set{\xstartfifthSeq}{\xstartfourthSeq + \labelSpace}
	\node [circle, draw, inner sep=1.5pt, fill] at (\xstartfifthSeq, \ystart) (fifthCircle) {};
	\node [anchor=north, below= 0.1cm of fifthCircle] (seqfourtext) {Seq $4$};
	
	\set{\xstartsixthSeq}{\xstartfifthSeq + \labelSpace}
	\node [circle, draw, inner sep=1.5pt, fill] at (\xstartsixthSeq, \ystart) (sixthCircle) {};
	\node [anchor=north, below= 0.1cm of sixthCircle] (seqfivetext) {Seq $5$};
	
	\set{\xstartsevenSeq}{\xstartsixthSeq + \labelSpace}
	\node [circle, draw, inner sep=1.5pt, fill] at (\xstartsevenSeq, \ystart) (seventhCircle) {};
	\node [anchor=north, below= 0.1cm of seventhCircle] (seqsixtext) {Seq $6$};
	
	\set{\xstarteightSeq}{\xstartsevenSeq + \labelSpace}
	\node [circle, draw, inner sep=1.5pt, fill] at (\xstarteightSeq, \ystart) (eigthCircle) {};
	\node [anchor=north, below= 0.1cm of eigthCircle] (seqseventext) {Seq $7$};
	
	\set{\xstartnineSeq}{\xstarteightSeq + \labelSpace}
	\node [circle, draw, inner sep=1.5pt, fill] at (\xstartnineSeq, \ystart) (ninethCircle) {};
	\node [anchor=north, below= 0.1cm of ninethCircle] (seqeighttext) {Seq $8$};
	
	\set{\xstarttenSeq}{\xstartnineSeq + \labelSpace}
	\node [circle, draw, inner sep=1.5pt, fill] at (\xstarttenSeq, \ystart) (tenthCircle) {};
	\node [anchor=north, below= 0.1cm of tenthCircle] (seqninetext) {Seq $9$};
	
	\set{\xstartelevenSeq}{\xstarttenSeq + \labelSpace}
	\node [circle, draw, inner sep=1.5pt, fill] at (\xstartelevenSeq, \ystart) (eleventhCircle) {};
	\node [anchor=north, below= 0.1cm of eleventhCircle] (seqtentext) {Seq $10$};
	
	\set{\xstarttwelfeSeq}{\xstartelevenSeq + \labelSpace}
	\node [circle, draw, inner sep=1.5pt, fill] at (\xstarttwelfeSeq, \ystart) (twelvethCircle) {};
	\node [anchor=north, below= 0.1cm of twelvethCircle] (seqeleventext) {Seq $11$};

	\node [anchor=north, right= 0.1cm of seqeleventext] {$\hdots$};

	\node [imgStyle, inner sep=0pt, above= 0.15cm of firstCircle, anchor=south] (firstImgFirst)  {\includegraphics[width=\imageSize cm]{pictures/new_classes/class_3.jpg}};
	\node [imgStyle, inner sep=0pt, above=0.1cm of firstImgFirst, anchor=south] (firstImgSecond) {\includegraphics[width=\imageSize cm]{pictures/new_classes/class_1.jpg}};
	\node [imgStyle, inner sep=0pt, above= 0.1cm of firstImgSecond, anchor=south] (firstImgThird) {\includegraphics[width=\imageSize cm]{pictures/new_classes/class_2.jpg}};
	
	\node [imgStyle, inner sep=0pt, above= 0.6cm of secondCircle, anchor=south] (firstImgFourth) {\includegraphics[width=\imageSize cm]{pictures/new_classes/class_0.jpg}};
	\node [imgStyle, inner sep=0pt, above= 0.1cm of firstImgFourth, anchor=south] (firstImgFivth) {\includegraphics[width=\imageSize cm]{pictures/new_classes/class_4.jpg}};
	
	\node [imgStyle, inner sep=0pt, above= 0.6cm of thirdCircle, anchor=south] (thirdImgFirst) 
	{\includegraphics[width=\imageSize cm]{pictures/new_classes/class_5.jpg}};
	\node [imgStyle, inner sep=0pt, above= 0.1cm of thirdImgFirst, anchor=south] (firstImgSecond) {\includegraphics[width=\imageSize cm]{pictures/new_classes/class_6.jpg}};
	
	\node [imgStyle, inner sep=0pt, above= 0.6cm of fourthCircle, anchor=south] (fourthImgFirst)
	{\includegraphics[width=\imageSize cm]{pictures/new_classes/class_7.jpg}};
	\node [imgStyle, inner sep=0pt, above= 0.1cm of fourthImgFirst, anchor=south] (firstImgFourth) {\includegraphics[width=\imageSize cm]{pictures/new_classes/class_8.jpg}};
	
	\node [imgStyle, inner sep=0pt, above= 0.6cm of fifthCircle, anchor=south] (fifthImgFirst) 	{\includegraphics[width=\imageSize cm]{pictures/new_classes/class_10.jpg}};
	\node [imgStyle, inner sep=0pt, above= 0.1cm of fifthImgFirst, anchor=south] (firstImgFivth) {\includegraphics[width=\imageSize cm]{pictures/new_classes/class_9.jpg}};	
	
	\node [imgStyle, inner sep=0pt, above= 0.6 of sixthCircle, anchor=south] (firstImgSecond) {\includegraphics[width=\imageSize cm]{pictures/new_classes/class_11.jpg}};
	\node [imgStyle, inner sep=0pt, above= 0.1cm of firstImgSecond, anchor=south] (firstImgThird) {\includegraphics[width=\imageSize cm]{pictures/new_classes/class_12.jpg}};
	
	\node [imgStyle, inner sep=0pt, above= 0.6cm of seventhCircle, anchor=south] (firstImgFourth) {\includegraphics[width=\imageSize cm]{pictures/new_classes/class_13.jpg}};
	\node [imgStyle, inner sep=0pt, above= 0.1cm of firstImgFourth, anchor=south] (firstImgFivth) {\includegraphics[width=\imageSize cm]{pictures/new_classes/class_14.jpg}};
	
	\node [imgStyle, inner sep=0pt, above= 0.6cm of eigthCircle, anchor=south] (firstImgFourth)	{\includegraphics[width=\imageSize cm]{pictures/new_classes/class_15.jpg}};
	\node [imgStyle, inner sep=0pt, above= 0.1cm of firstImgFourth, anchor=south] (firstImgSecond) {\includegraphics[width=\imageSize cm]{pictures/new_classes/class_16.jpg}};
	
	\node [imgStyle, inner sep=0pt, above= 0.6cm  of ninethCircle, anchor=south] (firstImgThird) {\includegraphics[width=\imageSize cm]{pictures/new_classes/class_17.jpg}};
	\node [imgStyle, inner sep=0pt, above= 0.1cm of firstImgThird, anchor=south] (firstImgFourth) {\includegraphics[width=\imageSize cm]{pictures/new_classes/class_18.jpg}};
	
	\node [imgStyle, inner sep=0pt, above= 0.6cm of tenthCircle, anchor=south] (firstImgFivth) {\includegraphics[width=\imageSize cm]{pictures/new_classes/class_19.jpg}};
	\node [imgStyle, inner sep=0pt, above= 0.1cm of firstImgFivth, anchor=south] (firstImgFivth)
	{\includegraphics[width=\imageSize cm]{pictures/new_classes/class_20.jpg}};

	\node [imgStyle, inner sep=0pt, above= 0.6cm and 0.05cm of eleventhCircle, anchor=south] (firstImgSecond) {\includegraphics[width=\imageSize cm]{pictures/new_classes/class_21.jpg}};
	\node [imgStyle, inner sep=0pt, above= 0.1cm of firstImgSecond, anchor=south] (firstImgThird) {\includegraphics[width=\imageSize cm]{pictures/new_classes/class_22.jpg}};

	\node [imgStyle, inner sep=0pt, above= 0.6cm of twelvethCircle, anchor=south] (firstImgFourth) {\includegraphics[width=\imageSize cm]{pictures/new_classes/class_23.jpg}};
	\node [imgStyle, inner sep=0pt, above= 0.1cm of firstImgFourth, anchor=south] (firstImgFivth) {\includegraphics[width=\imageSize cm]{pictures/new_classes/class_24.jpg}};
	
	\end{tikzpicture}
\caption{Long version of \DLRDataset dataset with twelve sequences. A particular set of categories is introduced to the network in each sequence, whereas each category contains a set of instances. For example, in the first sequence, the categories camera, ball, and bowl are introduced. Different instances of the same category are used for testing, and these categories are only available in the first sequence.}
\label{fig:howsLong}
\end{figure*}